\documentclass[conference]{IEEEtran}

\usepackage{xcolor}
\usepackage{soul}

\sethlcolor{yellow}

\usepackage{todonotes} 
\let\oldtodo\todo
\renewcommand{\todo}[1]{\oldtodo[inline]{#1}}

\usepackage{cite}
\usepackage{amsmath,amssymb,amsfonts}
\usepackage{graphicx}
\usepackage{textcomp}
\usepackage{xcolor}

\usepackage{color,soul} 

\usepackage{booktabs}
\usepackage{multirow}

\usepackage{algorithm}
\usepackage{algpseudocode}

\usepackage{siunitx}

\usepackage{makecell}

\usepackage{blindtext}

\usepackage{comment}

\usepackage{graphicx} 

\usepackage{amssymb}
\usepackage{listings} 
\usepackage{verbatim}
\usepackage{booktabs}
\usepackage{float}

\usepackage{tabularx} 
\usepackage{xcolor}

\usepackage{amsmath}

\usepackage{amsmath,amssymb,amsfonts}
\usepackage{graphicx}
\usepackage{textcomp}

\IEEEoverridecommandlockouts

\begin{document}


\title{ArrythML-II: An Optimized TinyML Framework for ECG Arrhythmia Detection and Multi-Class Beat Classification on Wearable Devices}



\author{
    Nagarajan S,
    and Kurian Polachan,~\IEEEmembership{Senior Member,~IEEE}%
    \\
    \textit{International Institute of Information Technology, Bangalore, India} \\
    \{nagarajan.s,  kurian.polachan\}@iiitb.ac.in

\thanks{An earlier version of this paper was presented at the IEEE 8th International Conference on Smart Applications, Communications and Networking (SmartNets), Rome, Italy, 2026 \cite{Nagarajan2026ECGSense}}
}

\maketitle

\begin{abstract}
This paper presents ArrythML-II, an optimized TinyML framework for on-device ECG arrhythmia detection and multi-class beat classification on wearable platforms. A systematic hyperparameter optimization framework is developed to identify compact autoencoder architectures for binary ECG arrhythmia detection. A custom scoring mechanism jointly considers detection performance, reconstruction-error thresholds, and model size to guide architecture selection. The framework is further extended to six-class ECG beat classification using a lightweight classifier derived from the optimized autoencoder architecture. Both models are INT8 quantized and deployed using TensorFlow Lite Micro on a custom-developed, battery-operated, single-lead ECG device based on the PSoC~6 microcontroller. The deployed autoencoder and classifier have model sizes of approximately 44~KB and 22~KB, respectively, with end-to-end processing times of 5.1~ms and 4.7~ms per ECG segment. Model inference accounts for 3.95~ms and 1.38~ms of these processing times, respectively. In the refined arrhythmia detection evaluation using the MIT-BIH Arrhythmia Database, the autoencoder achieved an accuracy of 84.76\%, precision of 82.68\%, and recall of 77.99\%.  The multi-class classifier achieved 82.29\% accuracy, weighted precision of 90.18\%, and weighted recall of 82.29\%. Beyond offline model evaluation, the wearable platform is validated using preloaded ECG segments, live ECG signals, and an ESP32-S3-based ECG emulator that replays annotated MIT-BIH recordings as analog waveforms. The results demonstrate the feasibility of integrating optimized TinyML models into wearable platforms for real-time ECG analysis.
\end{abstract}

\begin{IEEEkeywords}
TinyML, ECG Analysis, Wearable Devices
\end{IEEEkeywords}

\section{Introduction}

An electrocardiogram (ECG) captures the electrical activity of the heart and is a widely used tool for monitoring and diagnosing cardiac abnormalities, including arrhythmias \cite{Moody2001}. Continuous ECG monitoring is particularly important for the early detection of abnormal heart rhythms, enabling timely medical intervention and reducing the risk of severe cardiovascular events \cite{kakria2015realtime}. Despite its clinical importance, conventional ECG analysis remains largely dependent on hospital-based equipment, manual inspection, and trained medical personnel, which can lead to delays in diagnosis and limit the feasibility of long-term monitoring \cite{Rangayyan2002}.

Wearable ECG monitoring devices offer a promising solution. In recent years, they have enabled cardiac monitoring outside clinical environments \cite{baig2017systematic}. These devices typically acquire ECG signals over extended periods and transmit the recorded data to a nearby hub or cloud server for post-processing and analysis, often using machine learning (ML) algorithms to detect abnormalities. While this approach improves accessibility, it introduces several challenges. Frequent wireless transmission of ECG data consumes significant energy, particularly when using communication technologies such as BLE or WiFi, thereby reducing battery life \cite{agarwal2024vlc}. In addition, transmitting sensitive physiological data raises privacy and security concerns, as intercepted signals may expose sensitive health information \cite{majumdar2025hbclock, kambhampati2025murmur}. Although encryption techniques can mitigate these risks, implementing secure key management and cryptographic operations further increases system complexity, energy consumption, and latency \cite{vaidya2020gpio, chakrabarty2024iopuf, naz2024led, porlapothula2025cappuf}. Moreover, reliance on off-device processing can delay anomaly detection, limiting the responsiveness of such systems.

An alternative approach is to perform ECG signal processing and abnormality detection directly on the wearable device itself\cite{warden2019tinyml,Heydari2025TinyMLSurvey,elhanashi2024advancements}. By executing ML algorithms locally, the need for frequent wireless transmission can be reduced, resulting in lower energy consumption, improved data privacy, and faster response times. However, realizing this approach on wearable hardware is challenging because many wearable devices are built around low-cost microcontroller units (MCUs). While MCUs are energy efficient and well suited for battery-powered operation, they are severely constrained in terms of processing capability and on-chip memory. These constraints make it difficult to deploy conventional ML models, which typically require substantial computational and memory resources for ECG classification tasks. Tiny Machine Learning (TinyML) offers a practical alternative by enabling machine learning models to operate within the computational and memory constraints of MCUs.

Our earlier work, ArrythML-I~\cite{Nagarajan2026ArrythML}, investigated the feasibility of applying TinyML techniques to ECG arrhythmia detection using an autoencoder-based model designed to operate within the flash and RAM constraints of MCUs. 
Although the results demonstrated the potential of TinyML for on-device ECG analysis, several aspects of model development and practical wearable deployment remained unexplored, as discussed in detail in Section~\ref{sec_system_design}.

To address these limitations, this work presents ArrythML-II, an extension of ArrythML-I that advances on-device ECG analysis through systematic model optimization, six-class ECG beat classification, and hardware-level evaluation using a custom-developed wearable. The main contributions are summarized as follows:

\begin{itemize}
    \item A systematic hyperparameter optimization framework is proposed to identify compact autoencoder architectures for arrhythmia detection. A custom scoring mechanism jointly considers detection performance, reconstruction-error thresholds, and model size to select architectures suitable for embedded deployment.

    \item Beyond arrhythmia detection, this work presents a lightweight classifier for six-class ECG beat classification built from the hyperparameter-optimized autoencoder architecture.

    \item A custom-developed, battery-operated wearable ECG device based on the PSoC-6 microcontroller is presented for single-lead ECG acquisition, embedded signal processing, and on-device TinyML inference. The deployed models are evaluated on this device using preloaded ECG segments and live ECG signals.

    \item Furthermore, an ESP32-S3-based ECG emulator is developed to replay annotated MIT-BIH recordings as analog waveforms to evaluate the  ECG acquisition and inference pipeline of the wearable device.
\end{itemize}

\subsection{Outline}
The remainder of this paper is organized as follows. Section~\ref{sec_related_work} reviews existing wearable ECG devices and approaches to on-device ECG analysis. Section~\ref{sec_system_design} presents the proposed on-device ECG analysis system, the development and hyperparameter optimization of the autoencoder and classification models, their INT8 quantization and embedded deployment, and the ECG emulator-based evaluation framework. Section~\ref{sec_results} presents the experimental results, including off-device model performance evaluation and hardware-level evaluation using preloaded ECG segments, live ECG signals, and emulator-generated waveforms. Finally, Section~\ref{sec_conclusion} summarizes the work and discusses directions for future work.

\section{Related Work}
\label{sec_related_work}

This section first reviews existing wearable ECG devices and the methods they employ for ECG analysis. It then discusses recent developments in on-device neural-network-based ECG analysis. Finally, it reviews our earlier ArrythML study, which demonstrated the feasibility of an autoencoder-based approach for on-device arrhythmia detection, and compares it with the present work.

\subsection{Wearable ECG Devices}

Cardiac monitoring systems employ different sensing technologies, including electrocardiography (ECG), photoplethysmography (PPG), acoustic sensing, and radar-based monitoring. These modalities capture different physiological characteristics associated with cardiac activity. ECG measures the electrical activity of the heart, whereas PPG measures changes in peripheral blood volume associated with cardiac cycles. Radar-based systems detect subtle chest-wall movements associated with cardiac activity, while digital stethoscopes capture acoustic signals generated by the heart. Among these modalities, ECG is widely regarded as the gold standard for assessing cardiac electrical activity because it directly captures the electrical signals associated with cardiac excitation. The resulting ECG waveform provides detailed information about cardiac rhythm and electrical conduction, as reflected in its characteristic waveform morphology, including P waves, QRS complexes, and T waves. These features make ECG particularly valuable for identifying and analyzing cardiac arrhythmias~\cite{s25092848}.

Recently, wearable ECG devices have gained increasing attention for continuous cardiac monitoring owing to their compact form factor, portability, and ease of use. These devices are typically battery-operated and can be worn close to the chest, enabling continuous ECG acquisition over extended periods, ranging from several days to approximately one or two weeks, depending on the device design and battery capacity. Long-term ECG monitoring facilitates the observation of cardiac rhythm variations over time and increases the likelihood of capturing intermittent or transient arrhythmic events that may otherwise go undetected during conventional short-duration ECG recordings obtained during hospital visits. Such continuous monitoring can support the early identification of rhythm abnormalities and provide valuable information for subsequent clinical assessment and diagnosis.

Among wearable ECG devices, single-lead configurations are widely adopted, particularly in compact, chest-worn devices, owing to their relatively simple electrode arrangement and reduced hardware complexity. A single-lead ECG records the electrical potential difference between electrodes at two locations, providing one view of the heart's electrical activity. Although a single lead cannot provide the spatial information available from a standard 12-lead ECG, it preserves important electrical waveform characteristics useful for cardiac rhythm monitoring and beat-level analysis~\cite{s25092848}. Consequently, single-lead ECG provides a practical configuration for developing compact and low-power wearable cardiac monitoring systems.

Several studies have investigated wearable ECG monitoring systems based on single-lead configurations~\cite{kakria2015realtime, clark2018wearable, alimbayeva2024wearable, rajbhandary2022ecg, niu2023diagnostic, kwon2024telemonitoring, choi2024exercise, shin2026comparative, abdou2024accelerator}. Single-lead ECG technology has also been adopted in commercial wearable devices. For instance, the VitalPatch biosensor developed by VitalConnect~\cite{vitalconnect2026vitalpatch} and the Zio XT monitor developed by iRhythm Technologies~\cite{irhythm2025zioxt} are chest-worn devices designed for ECG monitoring over extended periods.

Although these systems facilitate continuous ECG acquisition, their capabilities for automated ECG interpretation vary considerably. Some primarily support ECG acquisition, recording, and wireless transmission, with the recorded waveforms subsequently reviewed by clinicians to identify potential abnormalities. Others incorporate automated arrhythmia detection or rhythm-analysis algorithms to identify abnormal cardiac events. In many implementations, ECG data are transmitted to a nearby computing device, such as a smartphone or gateway, or to a cloud platform for further analysis. The resulting diagnostic information is then made available to users or clinicians.

\subsection{On-Device ECG Analysis}

In many wearable ECG monitoring systems that support automated ECG analysis, analysis is performed on external computing platforms rather than directly on the wearable device. The acquired ECG signals are typically transmitted to a nearby computing device, such as a smartphone or gateway, or to a remote server for further processing and analysis. However, performing ECG analysis directly on the wearable device offers several potential advantages, including improved data privacy by minimizing the transmission of sensitive physiological information, reduced communication energy consumption by avoiding frequent wireless data transfers, and lower dependence on network connectivity. 

Conventional ECG analysis has traditionally used signal-processing techniques and rule-based methods to extract specific features from ECG waveforms for identifying cardiac abnormalities. These methods generally require less computational power and memory, making them suitable for implementation on resource-constrained microcontrollers. However, their performance depends on the features selected and the predefined detection rules. These parameters often require careful tuning to accommodate inter-subject variations in ECG morphology and differences in recording conditions. Furthermore, predefined features and rules may not adequately capture the diverse and complex waveform patterns associated with different cardiac abnormalities.

To overcome these limitations, machine learning (ML) and, more recently, deep learning (DL) approaches have been increasingly investigated for automated ECG analysis. Unlike conventional rule-based methods, these approaches can learn discriminative representations directly from ECG signals, reducing the dependence on manually designed features. Neural-network architectures, based on deep neural networks (DNNs) and convolutional neural networks (CNNs),  have been investigated for automated ECG classification~\cite{das2019heartbeatclassificationwearablesusing,ingolfsson2021ecgtcn,busia2024tinytransformer,ecgcnnclassification}. By learning relevant temporal and morphological characteristics from ECG waveforms, these architectures can capture complex patterns associated with different cardiac rhythms and support automated arrhythmia detection and classification. Among DL-based approaches, autoencoders are particularly compelling  architectures for ECG analysis {as they effectively capture complex cardiac morphology variations without hand-engineered features}. Unlike many DL-based approaches requiring labeled examples of different cardiac abnormalities, autoencoders can be trained primarily using normal ECG segments. This reduces dependence on labeled abnormal ECG data, which can be difficult to obtain in sufficient quantities~\cite{Nagarajan2025Autoencoder}. Furthermore, the autoencoder architecture enables the examination of waveform characteristics that contribute to the classification of an ECG segment as normal or abnormal, as discussed in Section~\ref{sec_system_design}. This capability to inspect the basis of a model's prediction is particularly valuable in healthcare applications, where root-cause analysis of diagnostic decisions is essential.

However, neural-network-based ECG analysis can require considerable memory and computational resources, depending on the model architecture and complexity, limiting its deployment on resource-constrained wearable devices built around MCUs. One approach to addressing these limitations is to develop lightweight neural-network architectures and apply optimization techniques, such as model compression and INT8 quantization, to reduce memory requirements and computational complexity while maintaining classification performance. 

Several studies have demonstrated lightweight neural-network-based ECG analysis on embedded platforms. Kim \textit{et al.}~\cite{Kim2023TinyMLECG} proposed a TinyML-based ECG classification system employing a compact CNN with a model size of approximately 27~KB, demonstrating its deployment on an Arduino-based embedded platform. Hizem \textit{et al.}~\cite{Hizem2025ECGEdge} proposed a CNN-based approach for ECG anomaly detection that incorporates preprocessing and feature extraction using heart rate variability (HRV). Their prototype used an Arduino Nano for ECG acquisition and a Raspberry Pi 4 for inference, with pruning and quantization applied to reduce model requirements. Akbari \textit{et al.}~\cite{Akbari2025AFEdge} demonstrated real-time atrial fibrillation (AF) detection on a low-power MCU using a compact bidirectional Long Short-Term Memory (LSTM) network, extracting HRV features from successive R--R intervals rather than processing ECG waveforms directly. More recently, Busia \textit{et al.}~\cite{busia2024tinytransformer} investigated a lightweight transformer architecture comprising approximately 6.6k parameters for five-class arrhythmia classification, deploying and evaluating the model on the multicore GAP9 processor.

Despite these advances, existing studies exhibit several limitations relative to the present work. First, although model compression techniques such as quantization and pruning have been explored, systematic investigation of network architectures and hyperparameter configurations remains limited. Second, existing evaluations do not consistently assess the complete ECG processing and inference pipeline under wearable operating conditions; while some works demonstrate embedded deployment or real-time signal acquisition, systematic stage-by-stage characterization, spanning R-peak detection, beat segmentation, and model inference using controlled ECG waveform replay, remains lacking. Third, prior implementations predominantly rely on commercial off-the-shelf (COTS) development platforms rather than custom-designed wearable ECG hardware. Furthermore, Busia \textit{et al.}~\cite{busia2024tinytransformer} demonstrate efficient transformer inference on the multicore GAP9 processor, which includes a dedicated AI accelerator. In comparison, the present work implements inference on the PSoC~6 Cortex-M4 CPU without dedicated AI acceleration.

Our earlier work, ArrythML-I~\cite{Nagarajan2026ArrythML}, investigated the feasibility of applying TinyML techniques to ECG arrhythmia detection using an autoencoder-based model designed to operate within the Flash and RAM constraints of microcontrollers (MCUs). In contrast, ArrythML-II builds upon this initial work and addresses several of its limitations.

First, ArrythML-I primarily focused on the proof-of-concept development and deployment of a TinyML model on an embedded platform. The model architecture was empirically selected and supported by detailed performance evaluation and failure analysis. In contrast, ArrythML-II adopts a systematic model design and hyperparameter optimization methodology, as described in Section~\ref{sec_system_design}, to identify suitable network architectures and configurations for resource-constrained deployment. Second, the evaluation in ArrythML-I relied on offline evaluation and on-device inference using pre-segmented ECG records from the MIT-BIH Arrhythmia Database stored on an SD card. ArrythML-II extends the evaluation to include both live ECG acquisition and analog waveform replay using an ECG emulator. Third, ArrythML-I focused exclusively on arrhythmia or anomaly detection. ArrythML-II extends this functionality by incorporating six-class ECG beat classification, distinguishing normal beats from different arrhythmic categories. This provides more detailed information about ECG beat types. Fourth, ArrythML-I evaluated the models using an ESP32-S3-based MCU development kit rather than a custom wearable ECG platform. In contrast, ArrythML-II presents the design and implementation of a custom single-lead ECG wearable, providing a more practical platform for evaluating TinyML models under wearable operating conditions.

\section{System Design}
\label{sec_system_design}

\begin{figure*}[!htbp]
    \centering
    \includegraphics[width=1\linewidth]
    {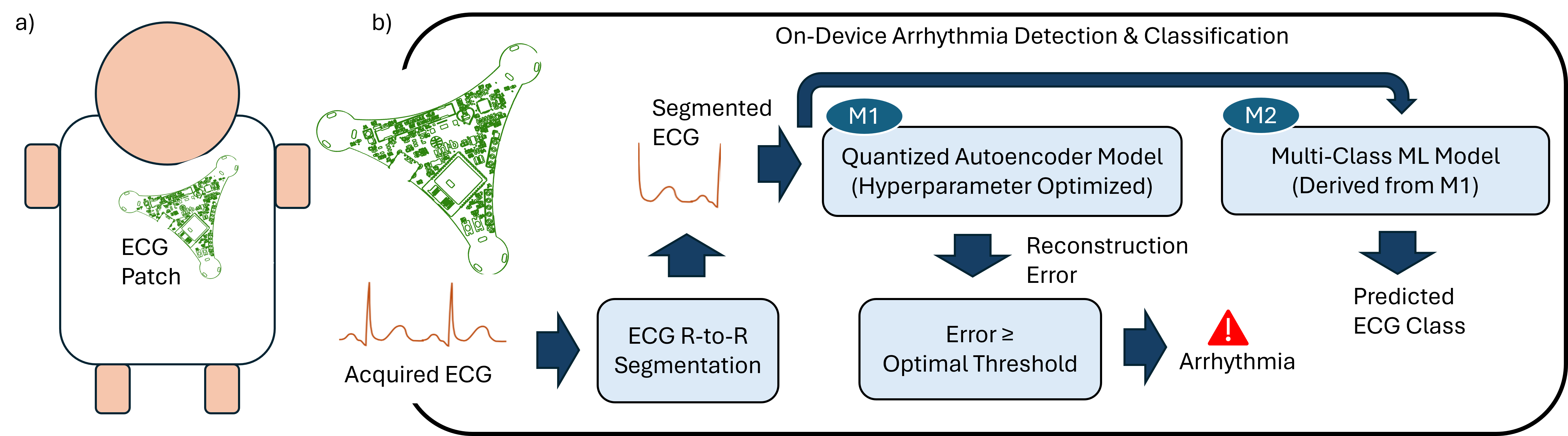}
\caption{Conceptual diagram of real-time, on-device arrhythmia detection and classification using TinyML models deployed on an ECG wearable. (a) Subject wearing the wearable; (b) On-device inference pipeline.}    \label{fig_system_design}
\end{figure*}

Fig.~\ref{fig_system_design} illustrates the proposed on-device ECG arrhythmia detection and arrhythmia classification framework. The system consists of a battery-operated, PSoC~6-based wearable ECG device that acquires a single-lead ECG signal from the user. The acquired ECG signal is periodically segmented and processed on the wearable, and the resulting ECG segments are analyzed by on-device TinyML models for arrhythmia detection and multi-class classification.

The remainder of this section describes the design of the major components of the proposed framework. In particular, it presents the TinyML model development for arrhythmia detection and classification, the hardware and firmware design of the wearable, and the evaluation framework, including the testbed developed to validate the performance of the models deployed on the wearable.

\subsection{On-Device Arrhythmia Detection and Classification}
\label{sec:on_device_detection_classification}

Fig.~\ref{fig_system_design} shows the proposed  wearable-based on-device arrhythmia detection and classification system. The ECG samples recorded by the  wearable are first post-processed to remove the DC offset and to center the ECG waveform around zero. Next, the R-peaks in the waveform are identified, and ECG segments are extracted between two consecutive R-peaks. Each extracted segment is then interpolated to a fixed number of samples and passed to one of the two pre-trained TinyML models for inference: an autoencoder-based model for arrhythmia detection and a multi-class model for arrhythmia classification.

\subsubsection{Autoencoder Model for Arrhythmia Detection}
\label{sec:autoencoder_arrhythmia_detection}

An autoencoder is a neural network composed of two interconnected components: an encoder and a decoder. The encoder compresses the input into a lower-dimensional latent representation, while the decoder reconstructs the original input from this latent representation.

In this work, the autoencoder is trained using normal ECG segments, with the objective of minimizing the reconstruction error between the input ECG segment and the corresponding segment reconstructed by the decoder. During inference, the reconstruction error is computed for each input ECG segment. Normal ECG segments are expected to produce relatively low reconstruction errors, whereas abnormal segments are expected to produce higher reconstruction errors. If the reconstruction error exceeds a predetermined threshold, the corresponding ECG segment is classified as abnormal and identified as an arrhythmia.

Autoencoder-based arrhythmia detection offers several advantages. First, it provides potential for interpretability. When an ECG segment is flagged as abnormal, the reconstructed segment generated by the decoder can be compared with the original input segment to identify differences in waveform morphology that contribute to the elevated reconstruction error. This comparison provides insight into the characteristics of the ECG segment that led to its classification as abnormal. Furthermore, the approach can detect abnormal ECG morphologies without requiring explicit training samples for each arrhythmia type, as the model is trained solely on normal ECG samples.

In this work, a DNN-based autoencoder is employed, with both the encoder and decoder constructed using fully connected neural network layers. Hyperparameter tuning is performed to identify a suitable network architecture that provides an appropriate balance between detection performance and model size.

The search space comprises encoder architectures with one, two, or three fully connected layers, with the number of neurons in each layer selected from 30 to 300 in increments of 30. For each candidate encoder architecture, a symmetric decoder is constructed with layer dimensions mirroring those of the encoder to reconstruct the input ECG segment. In addition to the network architecture, the reconstruction-error threshold is treated as a hyperparameter. Thresholds corresponding to the 85th, 95th, and 98th percentiles of the reconstruction-error distribution obtained from the training data are evaluated. For hyperparameter tuning, 25 encoder architectures were randomly selected from the search space, and all three reconstruction-error thresholds were evaluated for each selected architecture, resulting in a total of 75 architecture–threshold evaluations.

For hyperparameter tuning, a training dataset comprising a mixture of normal and abnormal ECG segments is used. The dataset is divided into training and validation subsets using an 80:20 split. Only the normal ECG segments from the training subset are used to train the selected autoencoder configurations. Each configuration is trained using quantization-aware training (QAT), which simulates the numerical effects of INT8 quantization during training. INT8 quantization is used to reduce the model size and memory requirements for embedded deployment. For candidate evaluation, each trained model is converted into a quantized TensorFlow Lite representation. This representation is used to determine the model size and evaluate model performance using TensorFlow Lite inference. Following training, the candidate configurations are evaluated on the validation subset to obtain the corresponding arrhythmia-detection performance metrics, including accuracy and recall. The model size of each configuration is also recorded for comparison.

After determining the arrhythmia-detection performance metrics and model size for each selected configuration, a score-based ranking is employed to identify the best-performing model. The scoring function combines precision and F1-score with a penalty for larger model sizes. The objective is to prioritize arrhythmia-detection performance while favouring compact models suitable for embedded deployment. The score is defined as

\begin{equation}
\mathrm{Score}=0.5\times\mathrm{Precision}+0.5\times\mathrm{F1}-0.1\times\frac{\mathrm{Model\ Size\ (KB)}}{100}.
\label{eq:tuning_score}
\end{equation}

Here, higher values of precision and F1-score increase the score, whereas a larger model size decreases the score. 

During candidate generation, the encoder depth for each of the 25 candidate architectures was selected randomly from one-, two-, and three-layer configurations. The probability of selecting a one-layer encoder was set to 20\%, whereas the probabilities of selecting two- and three-layer encoders were each set to 40\%. Consequently, the randomized search was expected to generate approximately 5 one-layer, 10 two-layer, and 10 three-layer candidates. A fixed random seed was used to ensure reproducibility of the candidate architectures and the evaluation procedure.

\subsubsection{Multiclass Model for Arrhythmia Classification}
\label{sec:multiclass_arrhythmia_classification}

To classify ECG segments into individual arrhythmia categories, a TinyML-based multiclass classification model is developed. The model reuses the hyperparameter-optimized TinyML autoencoder architecture discussed in Section~\ref{sec:autoencoder_arrhythmia_detection}, with the following modification. The decoder of the autoencoder is replaced with a fully connected layer containing six neurons, with each neuron corresponding to one ECG beat category from the MIT-BIH convention: paced beat (\texttt{/}), atrial premature beat (\texttt{A}), 
left bundle branch block beat (\texttt{L}), normal sinus beat (\texttt{N}), 
right bundle branch block beat (\texttt{R}), and premature ventricular contraction (\texttt{V}). A softmax activation converts these scores into six normalized probabilities that add up to 100\%. Each probability represents the model's estimated likelihood that the input ECG segment belongs to the corresponding beat category. The category with the highest probability is selected as the predicted ECG beat class. The multiclass model is then trained separately using labeled ECG segments with quantization-aware training and subsequently converted into an INT8-quantized model for embedded deployment.

\subsection{Wearable System Design}

A custom single-lead wearable is designed to prototype the on-device arrhythmia detection framework shown in Fig.~\ref{fig_hw_design}. The patch supports real-time ECG signal acquisition, signal processing, and TinyML inference for arrhythmia detection and ECG beat classification. In this section, we first describe the hardware architecture of the wearable, followed by the embedded firmware design.

\subsubsection{Hardware Design}

\begin{figure}[!htbp]
    \centering
    \includegraphics[width=1\linewidth]{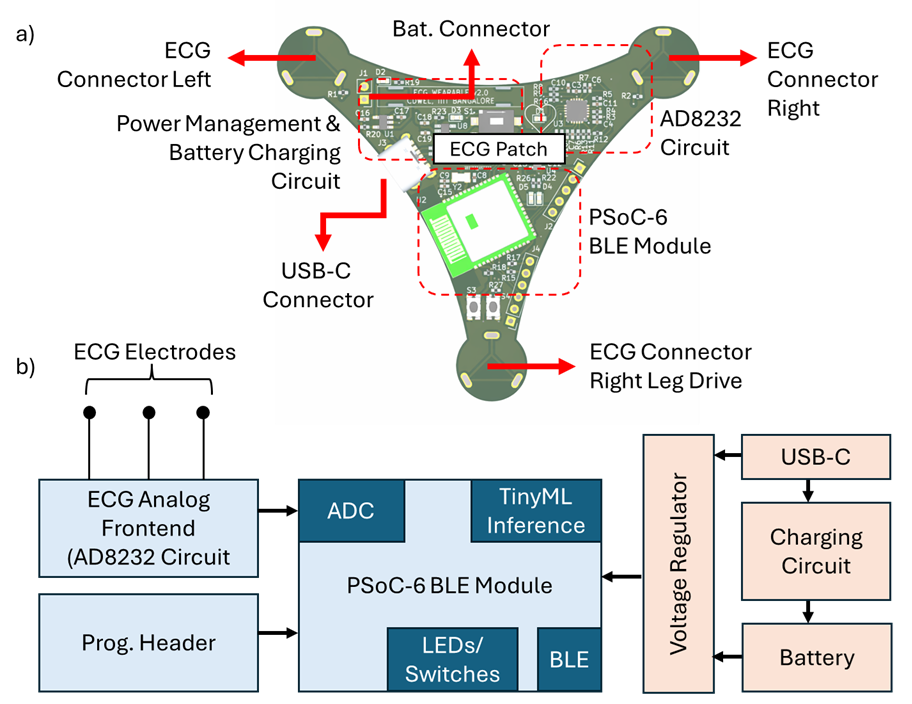}
\caption{(a) Developed wearable. (b) High-level hardware architecture of the proposed wearable ECG system.}
    \vspace{-3mm}
    \label{fig_hw_design}
\end{figure}

Fig~\ref{fig_hw_design}(a) presents the physical form factor of the wearable, and Fig.~\ref{fig_hw_design}(b) shows its block-level hardware architecture. The single-lead wearable is designed for real-time ECG acquisition and on-device TinyML inference. ECG signals are acquired using three surface electrodes (LA, RA, and RLD) placed on the chest and connected to the patch through snap connectors. The LA and RA electrodes acquire the differential ECG signal, while RLD provides the driven reference for common-mode noise reduction. The acquired signals are conditioned using an AD8232-based analog front-end (AFE)~\cite{ad8232_datasheet} configured for cardiac monitoring. The AFE provides signal amplification and biasing, band-pass filtering for noise suppression, and right-leg drive (RLD)-based common-mode noise reduction. The conditioned ECG signal is then sampled and digitized by the PSoC~6 microcontroller using its on-chip ADC~\cite{psoc6_proto_ble}, configured for 12-bit conversion over a 0--3.3~V input range.

The digitized ECG samples are subsequently processed locally by the PSoC~6 for R-peak detection, R--R interval segmentation, preprocessing, and TinyML inference for arrhythmia detection and beat classification. The onboard components are powered by a 3.7~V, 250~mAh lithium-polymer battery, which can be recharged through an external USB-C power source using a dedicated charging controller, MCP73831. The complete system is implemented on a compact custom PCB in a wearable form factor suitable for chest-mounted operation.

\subsubsection{Embedded Firmware Design}

Fig.~\ref{fig_system_design}(b) illustrates the embedded firmware pipeline implemented on the wearable. Following acquisition and analog conditioning by the AD8232-based AFE, the ECG signal is sampled and digitized using the on-chip ADC of the PSoC~6 microcontroller. ECG samples are acquired in consecutive five-second windows at a sampling rate of 360 samples/s. Each acquisition window therefore contains 1,800 samples, which are stored in a buffer for subsequent processing.

Once all samples in an acquisition window have been collected, the DC offset present in the acquired ECG waveform is estimated and removed from the samples. This offset correction centers the ECG waveform around zero and prepares the signal for subsequent R-peak detection.

Following offset correction, R peaks are identified using a custom local-maxima detection algorithm. A sample is considered a local maximum when its amplitude is greater than the preceding sample and greater than or equal to the following sample. A detected local maximum is accepted as an R peak only if it exceeds a predefined amplitude threshold and is separated from the previously accepted R peak by at least 0.5~s (180 samples). The amplitude threshold filters out low-amplitude signal variations that are unlikely to correspond to R peaks, while, the minimum separation prevents multiple R-peak detections within the same QRS complex.

The detected R peaks serve as reference points for ECG segmentation. The samples between each pair of consecutive R peaks are extracted as an R--R segment for further processing. Samples occurring before the first detected R peak and after the last detected R peak are excluded because they do not form complete R--R intervals. At least one complete R--R segment must be extracted from an acquisition window before the inference pipeline can proceed. Each extracted segment is linearly interpolated to a fixed length of 300 samples and independently normalized to the range $[0,1]$. The preprocessed segments are then provided to the deployed TinyML models for arrhythmia detection and ECG beat classification.

Algorithm~\ref{alg:ecg_tinyml} summarizes the complete embedded firmware flow, including ECG acquisition, DC-offset correction, R-peak detection, R--R interval segmentation, interpolation, normalization, and TinyML inference.

\paragraph*{Note}
If an R peak occurring between two successive cardiac cycles is missed, the interval between the two subsequently detected R peaks spans more than one actual R--R interval. Consequently, all samples between these two detected peaks are treated as a single segment. During autoencoder-based arrhythmia detection, the altered morphology of such a segment may produce a higher reconstruction error and, consequently, be identified as abnormal. The current implementation does not apply maximum R--R interval duration or physiological plausibility checks; such checks could be incorporated in future firmware revisions to reject unusually long segments before inference.

\begin{algorithm*}[!htbp]
\caption{Embedded firmware flow for ECG acquisition, preprocessing, and on-device arrhythmia detection.}
\label{alg:ecg_tinyml}

\begin{algorithmic}[1]
\Procedure{DetectArrhythmia}{}
    \State Initialize GPIO, UART, system clocks, and ADC
    \State Configure ADC sampling rate to $360$~Hz
    \State Initialize ECG buffer for a $5$~s acquisition window
    \State Load the INT8 TinyML model and detection threshold from Flash memory

    \While{device is active}

        \State $ecg \gets$ \Call{AcquireECG}{$1800$ samples}
        \State $ecg \gets$ \Call{RemoveDCOffset}{$ecg$}
        \State $peaks \gets$ \Call{DetectRPeaks}{$ecg$}

        \If{\Call{Length}{$peaks$} $< 2$}
            \State \textbf{continue}
        \EndIf

        \For{$i \gets 1$ to \Call{Length}{$peaks$}$-1$}

            \State $segment \gets$
            \Call{ExtractRRSegment}{$ecg, peaks[i], peaks[i+1]$}

            \State $segment \gets$
            \Call{LinearInterpolate}{$segment, 300$}

            \State $segment \gets$
            \Call{MinMaxNormalize}{$segment$}

            \State $error \gets$
            \Call{TinyMLInference}{$segment$}

            \If{$error > threshold$}
                \State \Call{ReportResult}{\textit{Abnormal}}
            \Else
                \State \Call{ReportResult}{\textit{Normal}}
            \EndIf

        \EndFor
    \EndWhile
\EndProcedure
\end{algorithmic}
\end{algorithm*}

\subsection{Evaluation Framework}
\label{sec:evaluation_framework}

Three approaches are adopted to evaluate the TinyML models: (1) offline evaluation, (2) on-device sample-based evaluation, and (3) on-device live and replay-based evaluation.

In the offline evaluation, the TinyML models are evaluated on a PC using the engineered datasets derived from the MIT--BIH Arrhythmia Database. Following offline evaluation, the trained models are deployed on the wearable for on-device sample-based evaluation using representative pre-recorded ECG samples stored in device memory. In this approach, the stored ECG samples are directly provided to the embedded firmware, bypassing the ECG signal acquisition hardware. This evaluation is used to validate the embedded ECG processing and TinyML inference pipeline using known and reproducible ECG inputs.

The replay-based evaluation extends the on-device evaluation a step further. In this approach, representative MIT--BIH ECG recordings are replayed as analog ECG signals using an ECG emulator connected to the wearable. The ECG signals are then acquired and digitized by the patch and subsequently processed by the embedded firmware for segmentation, pre-processing, and TinyML inference. Unlike sample-based evaluation, this approach also evaluates the analog ECG signal acquisition chain and its interaction with the embedded processing and inference pipeline. The replay-based evaluation framework and the ECG emulator developed for this purpose are described in this section.

\begin{figure}[!htbp]
    \centering
    \includegraphics[width=1\linewidth]{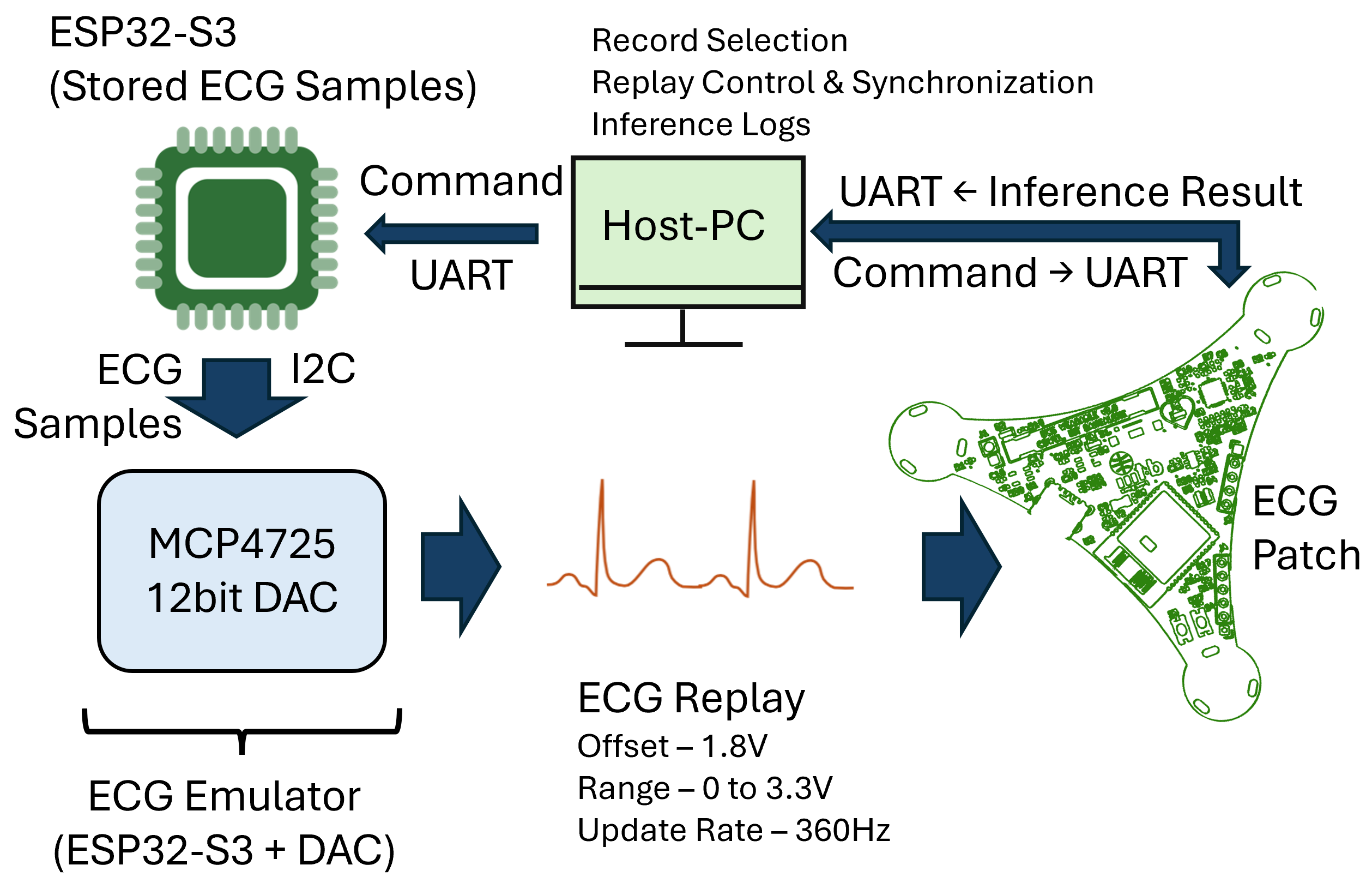}
    \caption{ Hardware setup for controlled ECG waveform replay and on-device evaluation. An ESP32-S3 replays stored MIT-BIH ECG samples via MCP4725 DAC to generate analog ECG waveforms to the wearable ECG patch. A host PC coordinates replay and subsequent evaluations over serial links.
}
    \label{fig_ReplayFrWork}
\end{figure}

Fig.~\ref{fig_ReplayFrWork} illustrates the replay-based evaluation framework, which comprises a host PC, an ECG emulator, and the wearable. The host PC coordinates the evaluation through serial communication with the ECG emulator and the wearable patch. The ECG emulator generates analog ECG waveforms from selected MIT--BIH recordings, which are subsequently acquired by the wearable patch for on-device processing.

The ECG emulator is implemented using an ESP32-S3 microcontroller and an MCP4725 12-bit digital-to-analog converter (DAC) interfaced through a Fast-mode I\textsuperscript{2}C interface operating at 400~kHz. Selected ECG excerpts from the ML-II channel of the MIT--BIH records are stored in the ESP32-S3 Flash memory as digital samples. During replay, the stored samples are read sequentially and converted to analog voltage samples through the MCP4725 DAC at the original MIT--BIH sampling rate of 360~Hz. The ECG amplitude is scaled and level-shifted with a 1.8~V bias to generate a conditioned waveform within the 0--3.3~V DAC output range.

During evaluation, the host-side PowerShell utility selects the required MIT--BIH record and coordinates waveform replay and ECG acquisition through serial commands. The emulator retrieves the corresponding ECG samples from Flash and generates the analog waveform at the original sampling rate of 360~Hz.

The AD8232 analog front-end is bypassed during replay-based evaluation, and the DAC output of the emulator is connected directly to the ADC input of the wearable patch. Since the MIT--BIH recordings represent low-amplitude ECG signals, direct reproduction of the original electrode-level millivolt waveform would require a higher-resolution, low-noise signal-generation path together with additional gain and amplitude calibration. Therefore, the emulator generates a conditioned waveform corresponding to the amplified and level-shifted signal expected at the output of the AD8232, rather than reproducing the original electrode-level ECG signal.

The wearable patch acquires the replayed waveform through its ADC and processes it using the same embedded pipeline used during normal operation, including ADC sampling, R-peak detection, R--R segmentation, interpolation, normalization, and TinyML inference. For each evaluation, the emulator replays a 10-s ECG waveform comprising 3,600 samples, while the wearable patch processes a five-second acquisition window. The additional replay duration provides sufficient time for synchronization between the emulator and the wearable patch. Following processing, the patch reports the detected R-peak locations, segment boundaries, inference results, and execution measurements through its serial interface.

The ECG emulator and wearable patch communicate independently with the host-side PowerShell utility through UART interfaces. Before each evaluation, the utility waits until both devices report their ready status. For a selected MIT--BIH record, the utility sends the record identifier to the emulator, which selects the corresponding waveform stored in Flash. The utility then initiates waveform replay and ECG acquisition on the wearable patch. After processing the acquisition window, the patch returns the results to the host, and the utility stores the corresponding evaluation logs.

The detected segment boundaries and inference outputs are subsequently compared with the corresponding MIT--BIH annotations to verify the operation of the embedded processing and inference pipeline. Since the replay framework evaluates a selected set of representative ECG excerpts, it is intended as a controlled hardware-level validation rather than a database-wide performance evaluation.

The overall replay procedure is summarized in Algorithm~\ref{alg:ecg_replay}.

\textbf{\begin{algorithm}[!htbp]
\caption{ECG Replay-Based Evaluation}
\label{alg:ecg_replay}
\begin{algorithmic}[1]
\State Open serial ports and evaluation log
\State Wait until both devices are ready
\For{each MIT--BIH record}
    \State Send record ID to emulator
    \State Wait until both devices are ready
    \State Start ECG replay and wearable acquisition
    \While{evaluation is ongoing}
        \State Collect replay status and inference results
        \State Log results
    \EndWhile
\EndFor
\State Close serial ports and log file
\end{algorithmic}
\end{algorithm}}

\section{Results}
\label{sec_results}

This section first presents the engineering of the datasets used for the development and evaluation of the TinyML models discussed in Section~\ref{sec:on_device_detection_classification}. It then presents the model training, hyperparameter selection, and offline performance evaluation of the autoencoder-based arrhythmia detection model, followed by the training and offline evaluation of the multi-class arrhythmia classifier. Subsequently, hardware-level evaluations are presented, including validation of ECG acquisition and R--R segmentation on the ECG wearable, followed by on-device evaluation of the firmware and TinyML models using representative ECG samples and replayed MIT--BIH ECG records.

\subsection{Dataset Engineering}
\label{sec_dataset_preparation}

The MIT--BIH Arrhythmia Database~\cite{MITBIHDatabase} was used to prepare the datasets for model development and evaluation of the TinyML models developed in this work. The database comprises 48 two-channel ambulatory ECG recordings, each approximately 30~min in duration, with expert-annotated beat locations and classifications. From this database, the modified Lead II (ML-II) channel was used to prepare the datasets. 

\subsubsection{R--R Segmentation and Labeling}

The datasets used for model development and evaluation were generated by extracting R--R ECG segments from the MIT--BIH recordings. Python scripts based on the WFDB library were developed to automate dataset preparation. 

For the development dataset, R peaks in the ECG waveforms were identified using the MIT--BIH annotations. Consecutive R peaks were then used to define individual R--R intervals. For the evaluation dataset, the ECG waveforms were first filtered using a Butterworth band-pass filter with cutoff frequencies of 0.5 and 50~Hz to attenuate baseline wander and high-frequency noise while retaining the dominant QRS frequency components required for R-peak detection. R peaks were subsequently identified using a custom peak-detection algorithm, and consecutive detected R peaks were used to extract R–R segments from the original, unfiltered ECG signal. This approach reflects the operation of the wearable under practical deployment conditions, where reference annotations are unavailable, and allows evaluation of the complete preprocessing and inference pipeline, including the effects of R-peak detection and R–R segmentation. In both cases, because the resulting R--R intervals have different durations, each extracted R--R segment was linearly interpolated to a fixed length of 300 samples and independently normalized to the range [0,1].

The ground-truth label assigned to an R--R segment depended on the ML task for which the dataset was used. For the autoencoder-based arrhythmia detection model, each segment was assigned a binary label, normal or abnormal. An R--R segment was labeled normal only when both R peaks bounding the segment were annotated as normal (N) in the MIT--BIH database. 

For the multi-class arrhythmia classifier, each R--R segment was assigned one of six target classes based on the MIT--BIH beat annotations: paced beat (/), atrial premature beat (A), left bundle branch block beat (L), normal beat (N), right bundle branch block beat (R), and premature ventricular contraction (V). In general, the class label was determined by the annotation associated with the starting (left) R peak of the segment. However, if an R--R segment contained a ventricular ectopic beat (V) within the corresponding R--R interval, the segment was labeled V irrespective of the annotation associated with its starting R peak.

\subsubsection{Development Dataset}
Two separate development datasets were prepared from the MIT-BIH Arrhythmia Database, with one used for the autoencoder-based arrhythmia detection model, requiring normal and abnormal labels for training and hyperparameter selection, and the other for the multi-class arrhythmia classifier, requiring six specific beat-class labels for model training.

For the autoencoder-based arrhythmia detection model, approximately 8,000 R--R ECG segments were extracted from the selected MIT--BIH records. Up to 200 normal and 200 abnormal segments were extracted from each selected record, subject to availability. During development dataset preparation, selected records were excluded due to deviations from typical normal ECG morphology (e.g., Record~\#233) or the presence of a large number of abnormal beats (e.g., Record~\#102) to reduce ambiguity in the normal class. The resulting dataset comprised approximately 5,000 normal and 3,000 abnormal segments.

The autoencoder development dataset was divided into an 80\% training subset and a 20\% held-out subset. The training subset contained approximately 4,000 normal and 2,400 abnormal segments, while the held-out subset contained approximately 1,000 normal and 600 abnormal segments. Only the normal segments in the training subset were used to train the candidate autoencoder models for hyperparameter selection. After training, both normal and abnormal segments in the held-out subset were used to evaluate the arrhythmia detection performance of each candidate autoencoder. The resulting performance was used to identify the best-performing architecture and reconstruction-error threshold, as described in Section~\ref{sec:ae_hyperparameter_tuning}.

Following the same R--R segment extraction procedure, a separate development dataset was prepared for the multi-class arrhythmia classifier. The dataset comprised approximately 8,700 R--R segments, of which approximately 58\% were normal segments, with the remaining segments distributed among the five abnormal classes. This dataset was used for model training.

\subsubsection{Evaluation Dataset}
\label{sec:evaluation_dataset}

A separate evaluation dataset was prepared from the MIT--BIH Arrhythmia Database for evaluating the developed models following the ECG segmentation and labeling procedures described above. ECG segments were extracted from the available MIT--BIH records, except for records~\#108, \#114, \#117, and \#203 which were excluded because of signal and annotation inconsistencies noted in these records~\cite{Nagarajan2026ArrythML}.
Approximately 95,000 R--R segments were extracted from the records to prepare the dataset. Among these, 93,000 segments corresponded to the six beat classes considered in this work while the remaining segments belonged to other, less frequently occurring annotation classes and were excluded from further evaluation. 

For multi-class evaluation, in the special case where a ventricular ectopic (V) annotation was present within an R–R segment or at either bounding R-peak, indicating ventricular ectopic morphology, the entire segment was assigned a V ground-truth label.

For the autoencoder, during inference verification, the refined binary evaluation assigned a normal ground-truth label to an R–R segment only when both bounding R-peaks were annotated as normal (N) and no abnormal beat occurred within the segment \cite{Nagarajan2025Autoencoder}. Otherwise, the segment was labeled abnormal. In an N–V–N pattern, where both bounding R-peaks are annotated as normal but an intermediate V beat occurs within the R–R segment, the ground truth was assigned as abnormal (V) rather than normal (N).

\subsection{Autoencoder-Based Arrhythmia Detection}
\label{sec:results_autoencoder_arrhythmia_detection}

\subsubsection{Hyperparameter Tuning and Model Selection}
\label{sec:ae_hyperparameter_tuning}
The optimal architecture for the autoencoder-based arrhythmia detection model was determined through hyperparameter tuning over a defined search space comprising 25 architectures and three reconstruction-error thresholds, resulting in a total of 75 architecture--threshold evaluations, as discussed in Section~\ref{sec:autoencoder_arrhythmia_detection}. The performance of each architecture--threshold combination was ranked using the score defined in Equation~\ref{eq:tuning_score}. The configuration achieving the highest score was selected as the optimal architecture for further evaluation and deployment.
The training dataset described in Section~\ref{sec_dataset_preparation}, comprising $\sim$8000 segments with $\sim$5000 normal and $\sim$3000 abnormal segments, was divided into training and test sets using an 80:20 split. The training set was used to train each of the 25 autoencoder architectures, while the corresponding test set was used to evaluate their performance and calculate the score.

\begin{figure}[!htbp]
\centering
\includegraphics[width=1\linewidth]{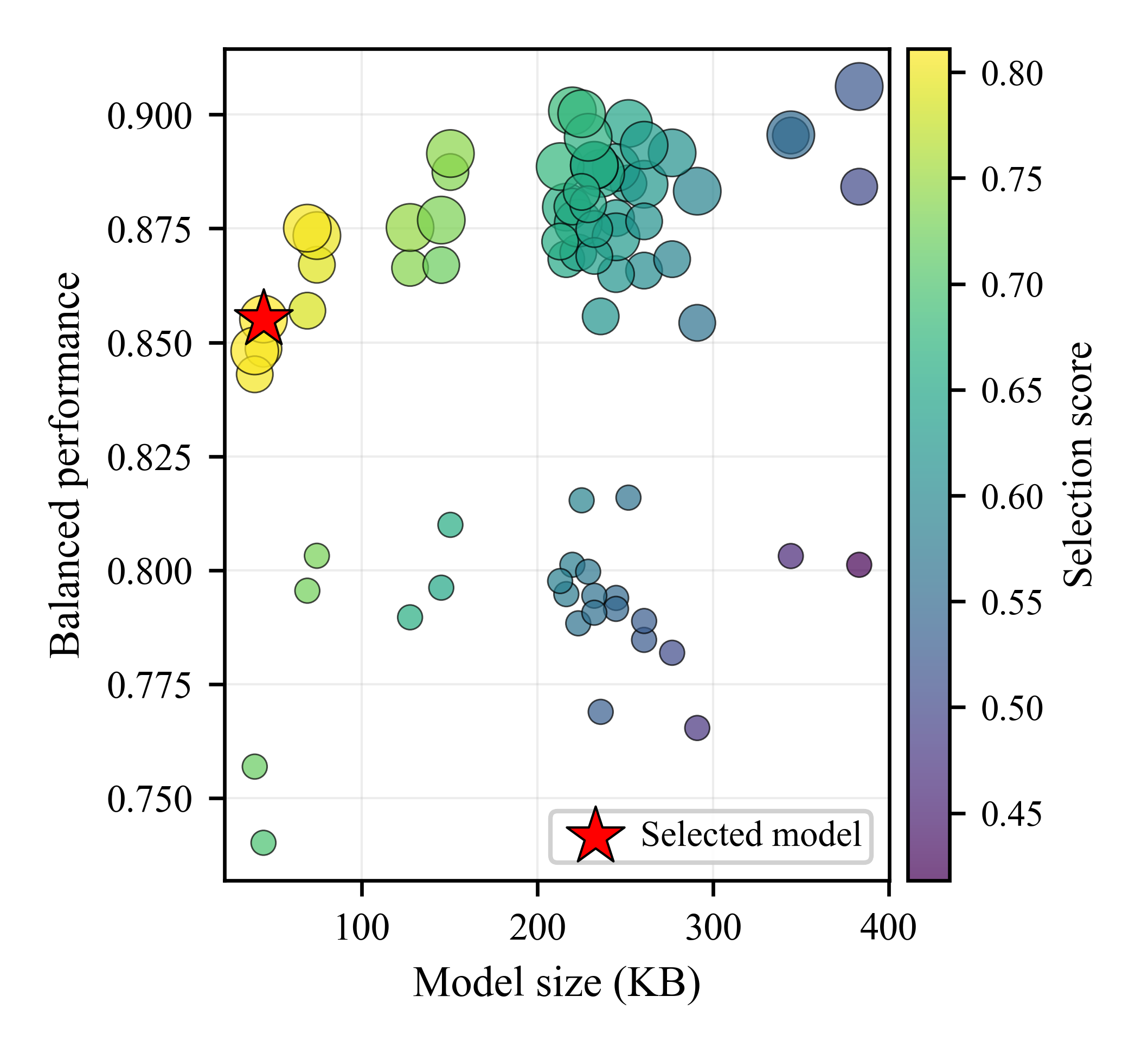}
\caption{DNN autoencoder hyperparameter search illustrating the trade-off between balanced performance and quantized model size. Marker size indicates the reconstruction-error threshold percentile, color indicates the selection score, and the star marks the selected configuration.}

\label{fig_hp_pareto}
\end{figure}

Fig.~\ref{fig_hp_pareto} presents the scores obtained from the 75 architecture--threshold evaluations. The figure illustrates the trade-off between balanced performance (the mean of precision and F1-score) and model size in selecting the appropriate model configuration. The ten configurations achieving the highest scores are listed in Table~\ref{tab_hp_results}, along with their architectural configurations, model sizes and performance metrics.

The model achieving the highest score was identified as a two-layer autoencoder with an encoder configuration of [60, 30] and a reconstruction-error threshold at the 98th percentile, denoted as [60, 30]p98. The [60, 30] configuration corresponds to encoder layers with 60 and 30 neurons, followed by a symmetric decoder with 30 and 60 neurons and a 300-neuron output layer to match the input segment dimension of 300. ReLU activation is used in the hidden layers of both the encoder and decoder, while sigmoid activation is applied to the output layer. The p98 designation indicates that the reconstruction-error threshold was set to the 98th percentile of the reconstruction-error distribution obtained during training. 

The results indicate that the selected two-layer encoder architecture outperformed both single-layer and three-layer encoder architectures in terms of the overall score. This finding demonstrates that, for resource-constrained embedded deployments, increasing model complexity does not necessarily result in improved performance and that an appropriate balance between detection performance and model size is essential.

\begin{table*}[!htbp]
\centering
\caption{Ten highest-scoring DNN autoencoder configurations from the hyperparameter search.}
\label{tab_hp_results}
\normalsize
\begin{tabular}{rlcccrrc}
\hline
\textbf{ID} & \textbf{Config} & \textbf{Layers} & \textbf{Percentile} &
\textbf{Accuracy} & \textbf{Recall} & \textbf{Size (KB)} & \textbf{Score} \\
\hline
1  & [60, 30] & 2 & 98 & 0.8641 & 0.6502 & 44.02 & 0.8111 \\
2  & [60]     & 1 & 98 & 0.8493 & 0.5883 & 39.05 & 0.8091 \\
3  & [90, 60] & 2 & 98 & 0.8752 & 0.6678 & 68.98 & 0.8061 \\
4  & [60, 30] & 2 & 95 & 0.8708 & 0.7014 & 44.02 & 0.8046 \\
5  & [60]     & 1 & 95 & 0.8696 & 0.7085 & 39.05 & 0.8040 \\
6  & [120]    & 1 & 98 & 0.8745 & 0.6678 & 74.41 & 0.7990 \\
7  & [120]    & 1 & 95 & 0.8893 & 0.7597 & 74.41 & 0.7926 \\
8  & [90, 60] & 2 & 95 & 0.8831 & 0.7509 & 68.98 & 0.7880 \\
9  & [120]    & 1 & 85 & 0.8752 & 0.9134 & 74.41 & 0.7288 \\
10 & [90, 60] & 2 & 85 & 0.8672 & 0.8852 & 68.98 & 0.7265 \\
\hline
\end{tabular}
\end{table*}

\subsubsection{Offline Evaluation and Deployment}

The selected autoencoder architecture was further evaluated using the evaluation dataset described in Section~\ref{sec_dataset_preparation}, comprising approximately 95,000 ECG segments. The evaluation followed the autoencoder-based arrhythmia detection method presented in our earlier work\cite{Nagarajan2026ArrythML}. In this method, each ECG segment is reconstructed using the trained autoencoder, and the reconstruction error between the input and reconstructed segments is computed. An ECG segment is classified as abnormal when its reconstruction error exceeds the threshold determined during hyperparameter tuning. The resulting predictions are then compared with the ground-truth labels to determine the classification performance. The ground truth for each R--R segment is determined from the MIT--BIH annotations of its bounding R-peaks. A segment is considered normal when both bounding R-peaks are annotated as normal. If the model also predicts the segment as normal, it is counted as a true negative (TN); otherwise, it is counted as a false positive (FP). Table~\ref{tab:perf_comparison} summarizes the evaluation metrics obtained on the evaluation dataset.

\begin{figure}[!htbp]
    \centering
    \includegraphics[width=1\linewidth]{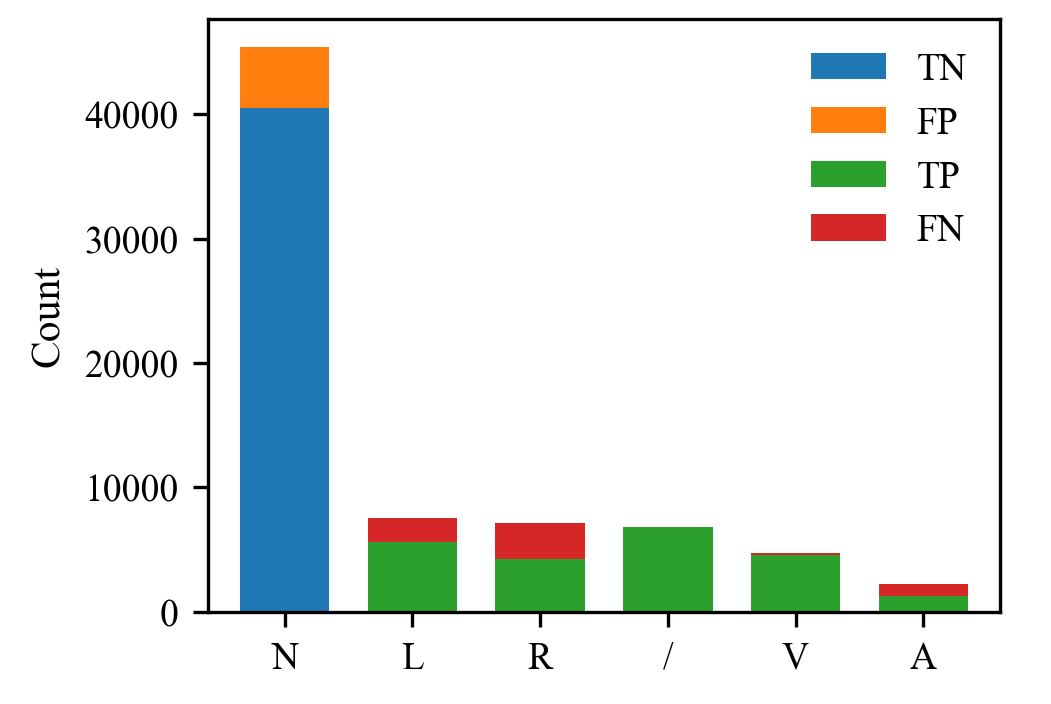}
\caption{Annotation-wise counts of true negatives, true positives, false positives, and false negatives for the DNN autoencoder. Only beat annotations with at least 500 samples are shown.}
    \label{fig_dnnAnnotMetrics}
\end{figure}

Fig.~\ref{fig_dnnAnnotMetrics} presents annotation-wise prediction outcomes of the quantized DNN autoencoder evaluated on the PC using the filtered MIT-BIH dataset. True positives (TP), true negatives (TN), false positives (FP), and false negatives (FN) are shown for each ECG annotation. Ambiguous beats and rhythm classes with fewer than 500 samples were excluded from the analysis. 

\begin{figure*}[!htbp]
    \centering
    \includegraphics[width=\textwidth]{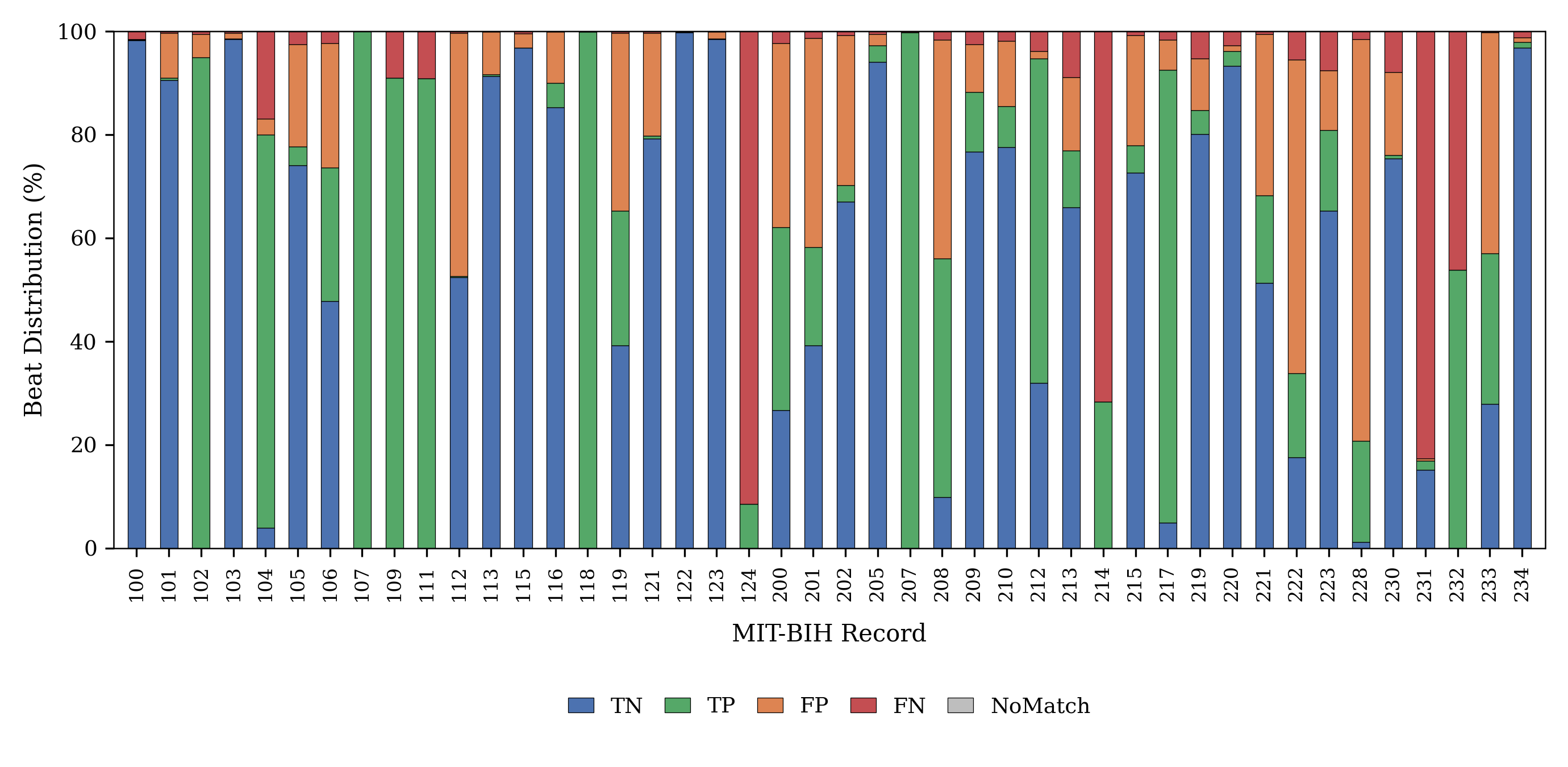}
    \caption{Record-wise percentages of true negatives, true positives, false positives, and false negatives for the DNN autoencoder on the unfiltered MIT--BIH evaluation dataset.}
    \label{fig_dnnRecordMetrics}
\end{figure*}

Fig~\ref{fig_dnnRecordMetrics} presents the record-wise prediction outcomes of the quantized DNN autoencoder evaluated offline using the MIT-BIH evaluation dataset. For each MIT-BIH record, the distribution of true negatives (TN), true positives (TP), false positives (FP), and false negatives (FN) is shown as a percentage of the evaluated ECG segments. The results illustrate how the detection outcomes vary across individual records and provide a record-level view of the model performance.

Following offline evaluation, the selected model was deployed on the target wearable hardware operating at a CPU frequency of 100~MHz. The model size was measured to be 44~KB. The on-device execution time was measured for both autoencoder inference and end-to-end ECG processing. The TensorFlow Lite Micro inference incurred a latency of approximately 3.95~ms per ECG segment. The end-to-end processing time, including ECG interpolation, normalization, autoencoder inference, reconstruction-error computation, and threshold-based classification, was approximately 5.1~ms per ECG segment.



\paragraph*{Refined Evaluation} For the autoencoder-based models, the default ground-truth assignment considered an R–R segment normal when both bounding R-peaks were annotated as normal (N); otherwise, the segment was labeled abnormal. However, this assignment could introduce false positives (FPs) in cases such as an N–V–N pattern, where a ventricular ectopic (V) beat occurs between two normal (N) beats. An R–R segment bounded by the two N beats could therefore be assigned normal ground truth despite containing ventricular ectopic morphology, resulting in an FP when the autoencoder predicted abnormal. Following the failure analysis reported in our earlier work, the refined evaluation reassessed only the FP cases associated with intermediate ectopic beats and morphology-related abnormalities \cite{Nagarajan2026ArrythML}. For N–V–N cases, segments containing an intermediate V beat were assigned abnormal ground truth. Morphology-related FP cases involving P-wave, ST-segment, or T-wave variations were excluded from the supplementary refined evaluation because their clinical significance could not be established using the available single-lead ML-II ECG configuration. False-negative (FN) cases were retained as model errors, reflecting limitations of the model and/or the single-lead ML-II ECG configuration. The refined evaluation results are presented in Table~\ref{tab:perf_comparison}.


\begin{table*}[!htbp]
\centering
\caption{Comparison of performance metrics for the INT8 DNN and CNN autoencoders and the multiclass ECG classifier.}
\label{tab:perf_comparison}
\renewcommand{\arraystretch}{1.2}
\begin{tabular}{|c|c|c|c|c|c|c|c|c|}
\hline
\textbf{Exp.} & \textbf{Evaluation} & \textbf{Model} & \textbf{Variant}
& \textbf{Accuracy (\%)} & \textbf{Precision (\%)} & \textbf{Recall (\%)}
& \textbf{F1-Score (\%)} & \textbf{Size (kB)} \\
\hline

\multirow{2}{*}{1}
& \multirow{2}{*}{Hyperparameter-tuned models}
& DNN-based & Quantized
& 76.45 & 63.70 & 77.58 & 69.96 & 44.0 \\
\cline{3-9}
&
& CNN-based & Quantized
& 77.97 & 64.91 & 82.04 & 72.48 & 18.0 \\
\hline

\multirow{2}{*}{2}
& \multirow{2}{*}{Refined evaluation}
& DNN-based & Quantized
& 84.76 & 82.68 & 77.99 & 80.27 & -- \\
\cline{3-9}
&
& CNN-based & Quantized
& 86.44 & 83.19 & 82.58 & 82.88 & -- \\
\hline

3
& Binary (mapped)
& Multi-class & Quantized
& 90.90 & 82.70 & 93.70 & 87.85 & 22.84 \\
\hline

4
& Multi-class classifier
& Multi-class & Quantized
& 82.29 & 90.18 & 82.29 & 83.11 & 22.84 \\
\hline
\end{tabular}
\end{table*}

\subsection{Multi-Class Arrhythmia Classifier}

\subsubsection{Model Construction and Training}
For the multi-class arrhythmia classifier, the optimized autoencoder model identified in Section~\ref{sec:results_autoencoder_arrhythmia_detection} is modified by replacing the decoder with a fully-connected classification layer as discussed in Section~\ref{sec:multiclass_arrhythmia_classification}.  The prepared training dataset discussed in Section~\ref{sec_dataset_preparation} is used for quantization-aware training (QAT) of this model.


The multi-class development dataset was first partitioned into an 80\% training pool and a 20\% held-out test subset. Within the training pool, 20\% was reserved for validation to monitor convergence and check for overfitting.

\subsubsection{Offline Evaluation and Deployment}
\label{sec:offline_evaluation}
\begin{figure}[!htbp]
    \centering
    \includegraphics[width=1\linewidth]{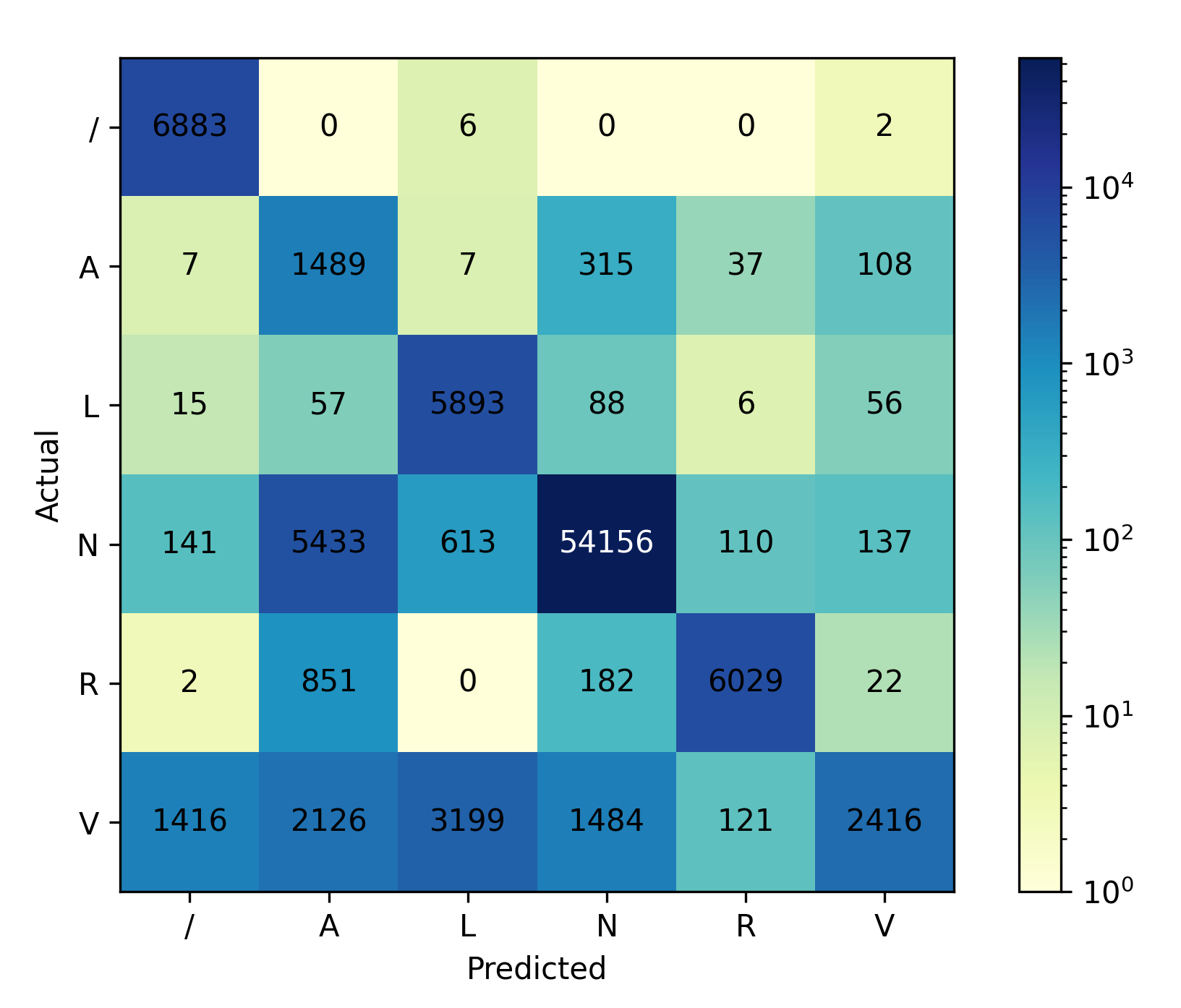}
\caption{Confusion matrix from PC-side evaluation of the INT8 TensorFlow Lite multiclass ECG beat classifier. Rows represent reference beat classes, and columns represent predicted beat classes.}
    \label{fig_conf_matrix}
\end{figure}

Post training, the INT8 quantized model was evaluated using the evaluation dataset in Section~\ref{sec:evaluation_dataset} on PC. Fig.~\ref{fig_conf_matrix} presents the resultant confusion matrix. The results indicate that the majority of ECG beats were correctly assigned to their corresponding rhythm categories. The confusion matrix shows that the majority of normal (N) beats were correctly classified as N. Among the misclassified beats, several errors occurred between abnormal rhythm classes rather than as abnormal-to-normal classifications. This indicates that, although the classifier may not always identify the exact arrhythmia subtype correctly, it can still distinguish a substantial proportion of abnormal beats from normal ECG activity.

The class-wise performance metrics demonstrate effective recognition of the evaluated ECG rhythm categories. The observed variations among the classes can partly be related to the information available from a single-lead ECG and individual R--R segments. For example, distinguishing the R and L classes can be limited when using a single ECG lead, while identification of A beats can also benefit from inter-beat variability information across consecutive beats. Such temporal information can be incorporated through additional software logic without modifying the deployed classifier. These factors help explain some of the observed class-wise variations and misclassifications.

Post-processing the confusion matrix yields the binary detection performance of the multi-class classifier model, which enables a direct comparison with the autoencoder model discussed in Section~\ref{sec:results_autoencoder_arrhythmia_detection}. In this evaluation, all ECG beats corresponding to normal rhythm were treated as normal samples, while all remaining rhythm categories were grouped into a single abnormal class. The multi-class predictions were subsequently mapped into binary outcomes, enabling direct comparison with earlier arrhythmia-detection results. Results are presented in Table~\ref{tab:perf_comparison}. The table also reports the multi-class classification performance.

Post offline evaluation, the model was deployed on the wearable hardware. The quantized model occupied approximately 22.8 KB of flash. The deployed implementation required approximately 1.38~ms for execution of the TensorFlow Lite Micro inference engine on the 100 MHz PSoC~6 platform. Including ECG segmentation, interpolation, normalization, and diagnostic instrumentation overhead, the complete segment-processing pipeline required approximately 4.7 ms per ECG segment.

\subsection{Hardware Evaluations}
\label{sec:results_hardware_evaluation}

\subsubsection{ECG Acquisition and Segmentation Validation}
\label{sec:results_ecg_acquisition_segmentation}

In this experiment, the ECG acquisition by the wearable and the subsequent R-peak detection performed by the firmware running on the wearable are validated.

\begin{figure*}[!htbp]
    \centering
    \includegraphics[width=0.8\linewidth]{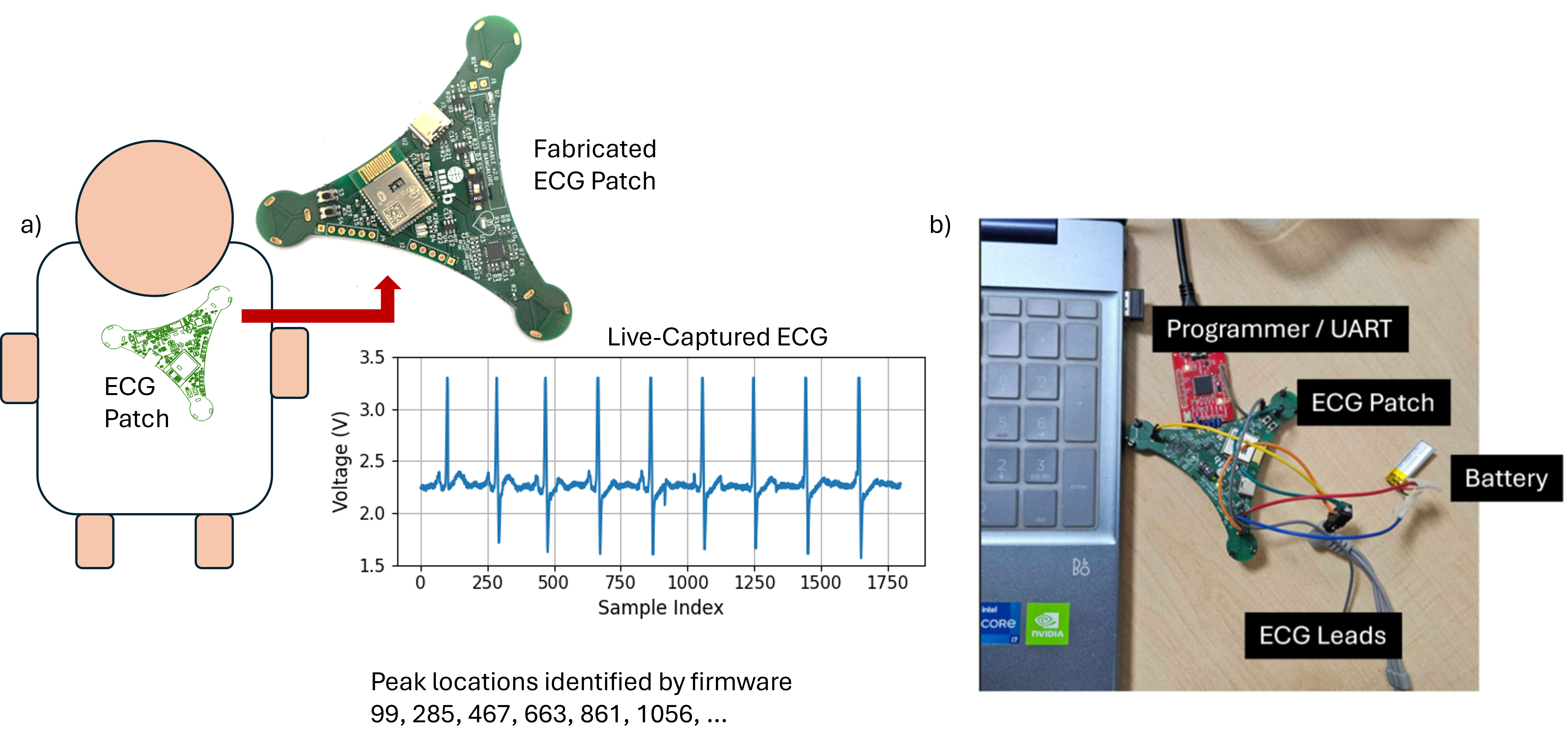}
\caption{Experimental setup for live ECG acquisition and on-device inference. (a) Experimental setup showing acquired ECG waveform. (b) Bench setup for testing the wearable.}
    \label{fig_hw}
\end{figure*}

For this experiment, the wearable shown in Fig.~\ref{fig_hw} is worn by a subject. ECG snap connectors connect the electrodes placed on the subject's chest to the wearable. Upon power-up, the wearable initializes its peripherals, performs integrity checks, and then starts acquiring the ECG signal. The on-board AD8232-based analog front-end (AFE) circuit first conditions the ECG signal and then passes it to the on-board PSoC~6. The signal is digitized using the PSoC~6 on-chip ADC. The digitized waveform is recorded and transferred to a PC for validation. Fig.~\ref{fig_hw}(a) shows a representative portion of the recorded waveform. The samples exhibit characteristic ECG features, such as QRS complexes and R-peaks, indicative of good-quality ECG acquisition from the subject. Furthermore, the ECG signal remains approximately centered around the mid-supply reference of the AD8232-based AFE, as expected, and is devoid of any saturation or clipping artifacts.

Alongside ECG acquisition, the wearable runs firmware to segment the acquired ECG. To verify the correctness of this operation, which primarily depends on accurately identifying the R-peak locations in the waveform, the R-peak locations identified by the firmware prior to R-to-R segmentation are recorded and compared with the corresponding peaks in the acquired waveform. Fig.~\ref{fig_hw}(a) shows the detected peak locations for a representative ECG recording. The detected locations closely correspond to the R-peaks in the waveform, as verified through visual inspection, confirming proper R-peak detection.

\subsubsection{On-Device TinyML  Evaluation}

In our prior work ~\cite{Nagarajan2026ArrythML}, quantized autoencoder models were evaluated using more than 95,000 ECG segments on the embedded platform and compared the results with the corresponding PC-side evaluations. The on-device and PC-side results showed close agreement in both model predictions and reconstruction-error thresholds, establishing the consistency of the inference results between the deployed TinyML models and their PC-side implementations. Based on this finding, we therefore do not repeat the extended offline evaluation presented in Section~\ref{sec:offline_evaluation} on the device. Instead, the present work evaluates the on-device implementation of the firmware and inference using representative ECG samples, both stored in memory and replayed using an emulator.

{For on-device evaluations, the TinyML models were deployed on the onboard PSoC~6 microcontroller of the wearable shown in Fig.~\ref{fig_hw}. The models were implemented using the TensorFlow Lite for Microcontrollers (TFLite Micro) framework through the vendor-provided \texttt{ml-tflite-micro} library within the ModusToolbox\textsuperscript{\texttrademark} development environment~\cite{InfineonModusToolboxSoftware}.
}

\paragraph{ECG Sample-Based Evaluation}

In this experiment, the firmware-based ECG processing pipeline and the autoencoder-based arrhythmia detection are co-validated using representative ECG samples. 

Representative samples from the MIT--BIH Arrhythmia Database, Record~100 representing the normal case and Record~107 representing the abnormal case, were stored in the PSoC~6 memory. The stored ECG samples were subsequently processed using the complete on-device pipeline, including R-peak detection, R--R-based segmentation, preprocessing, and autoencoder-based TinyML inference for arrhythmia detection.

Table~\ref{tab:tinyml_hw_results} summarizes the resulting segmentation and inference outputs. The \textit{Reference MIT--BIH Annotations} columns list the R-peak locations and corresponding annotations provided with the database for the selected ECG waveforms. The \textit{Segmentation and Inference on wearable} columns report the R--R segments identified by the firmware running on the wearable, the corresponding autoencoder reconstruction error, the number of processor clock cycles required for inference, the resulting inference latency, and the predicted class.

For Record~100, the firmware identifies the R-peak locations and forms the corresponding R--R segments, such as $(77,370)$, $(370,663)$, $(663,947)$, $(947,1231)$, and $(1231,1515)$. The detected segment boundaries closely match the R-peak locations provided in the MIT--BIH annotations. The reconstruction errors for these segments range from 0.0442 to 0.0630, and all segments are classified as \textit{Normal}. For Record~107, the identified R--R segment boundaries similarly correspond closely to the annotated R-peak locations. The resulting reconstruction errors are higher, ranging from 0.1062 to 0.1547, and all evaluated segments are classified as \textit{Abnormal}. Thus, the reconstruction-error-based classification correctly distinguishes the representative normal and abnormal ECG segments. Furthermore, the clock-cycle measurements for inference are reported in the table. At a PSoC~6 CPU clock frequency of 100~MHz, these measurements correspond to an inference latency of approximately 8.8~ms per ECG segment, which remains consistent across the evaluated segments. These results and inference timings are with baseline DNN autoencoder model~\cite{Nagarajan2026ECGSense}.

The \textit{Segmentation and Inference on PC-side} columns provide the corresponding results obtained using the Python implementation of the same processing pipeline. The segment boundaries, reconstruction errors, and predicted classes obtained on the wearable are in close agreement with the PC-side results for both records. This agreement verifies that the firmware implementation of the segmentation and the deployed autoencoder-based TinyML inference produce results consistent with the PC-based implementation.

In addition to the database-derived ECG samples, Table~\ref{tab:tinyml_hw_results} presents segmentation and inference results for a representative ECG snippet acquired directly from a subject using the wearable, following the experimental setup as shown in Fig.~\ref{fig_hw}(a). The firmware successfully identifies the R-peaks and forms the corresponding R--R segments. All evaluated segments are classified as \textit{Normal}.

Following the autoencoder evaluation, the multi-class classifier was evaluated on the wearable hardware to verify on-device classification across the selected ECG rhythm classes. Table ~\ref{tab:ecg_replay_validation} summarizes the deployment characteristics, timing measurements, and classification results obtained across different MIT-BIH rhythm categories on the wearable patch. Each row corresponds to a selected MIT-BIH record and a representative R–R segment derived from preloaded ECG samples and processed by the wearable patch. The segment boundaries indicate the ECG samples used for inference, while the MIT-BIH annotation provides the reference rhythm class. The autoencoder output shows whether the segment was detected as normal or abnormal, whereas the multi-class output provides the predicted rhythm class. The final two columns report the segment processing time and inference cycles on the embedded platform. Inference time represents the execution time of the TinyML model for a single ECG segment, whereas processing time includes preprocessing and minimal debug logging in addition to inference for each segment. Overall, the table demonstrates the operation of the models on preloaded ECG samples and their execution on the wearable hardware.

\paragraph{Note}
For experimental comparison, a CNN autoencoder was developed by reshaping each 300-sample ECG segment into a $15 \times 20 \times 1$ matrix, representing 15 rows of 20 samples with a single channel. The encoder comprises three convolutional layers with 32, 16, and 8 filters using $3 \times 3$ kernels and $2 \times 2$ max-pooling, while the decoder employs convolutional layers and $2 \times 2$ upsampling to reconstruct the original ECG segment. Quantization-aware training produced a TensorFlow Lite model of 17.56 KB. On the PSoC~6 platform operating at 100 MHz, the CNN achieved an average inference time of approximately 209 ms and a total processing time of approximately 210~ms per ECG segment. The classification performance is presented in Table~\ref{tab:perf_comparison}.

\begin{table*}[!htbp]
\centering
\caption{On-Device ECG Segmentation and TinyML inference Results, compared with PC-based reference results and ground-truth R-peak locations and annotations from the MIT--BIH database.}
\label{tab:tinyml_hw_results}
\renewcommand{\arraystretch}{1.15}
\setlength{\tabcolsep}{5pt}

\begin{tabular}{ccc | ccccc | ccc}
\hline
\multicolumn{3}{c|}{\textbf{Reference MIT--BIH Annotations}} &
\multicolumn{5}{c|}{\textbf{Segmentation and Inference on wearable}} &
\multicolumn{3}{c}{\textbf{Segmentation and Inference on PC-side}} \\

\hline
\textbf{Record} & \textbf{R-Peak} & \textbf{Anno.} &
\textbf{Segment} & \textbf{Re.Err.} & \textbf{Cycles} & \textbf{Time (ms)} & \textbf{Class} &
\textbf{Segment} & \textbf{Re.Err.} & \textbf{Class} \\
\hline

100 & 18   & + & (--,--)       & --     & --     & --    & --       & (--,--)       & --     & -- \\
100 & 77   & N & (--,--)       & --     & --     & --    & --       & (--,--)       & --     & -- \\

100 & 370  & N & (77,370)      & 0.0531 & 882369 & 8.8237 & Normal   & (77,370)      & 0.0529 & Normal \\
100 & 662  & N & (370,663)     & 0.0442 & 882018 & 8.8202 & Normal   & (370,663)     & 0.0441 & Normal \\
100 & 946  & N & (663,947)     & 0.0630 & 882170 & 8.8217 & Normal   & (663,947)     & 0.0632 & Normal \\
100 & 1231 & N & (947,1231)    & 0.0472 & 882936 & 8.8294 & Normal   & (947,1231)    & 0.0475 & Normal \\
100 & 1515 & N & (1231,1515)   & 0.0509 & 881753 & 8.8175 & Normal   & (1231,1515)   & 0.0518 & Normal \\

\hline

107 & 47   & + & (--,--)       & --     & --     & --    & --       & (--,--)       & --     & -- \\
107 & 95   & / & (--,--)       & --     & --     & --    & --       & (--,--)       & --     & -- \\

107 & 272  & / & (85,268)      & 0.1062 & 880916 & 8.8090 & Abnormal & (51,269)      & 0.1651 & Abnormal \\
107 & 569  & / & (268,565)     & 0.1529 & 884798 & 8.8480 & Abnormal & (269,566)     & 0.1444 & Abnormal \\
107 & 884  & / & (565,881)     & 0.1547 & 884690 & 8.8470 & Abnormal & (566,882)     & 0.1435 & Abnormal \\
107 & 1206 & / & (881,1202)    & 0.1389 & 884702 & 8.8470 & Abnormal & (882,1203)    & 0.1360 & Abnormal \\
107 & 1519 & / & (1202,1515)   & 0.1378 & 884581 & 8.8460 & Abnormal & (1203,1516)   & 0.1221 & Abnormal \\

\hline
\hline

Live-Captured & 99   & -- & (--,--)       & --     & --     & --     & --     & (--,--) & -- & -- \\

 & 285  & -- & (99,285)      & 0.0396 & 880512 & 8.8051 & Normal & -- & -- & -- \\
 & 467  & -- & (285,467)     & 0.0583 & 883324 & 8.8332 & Normal & -- & -- & -- \\
 & 663  & -- & (467,663)     & 0.0441 & 883339 & 8.8334 & Normal & -- & -- & -- \\
 & 861  & -- & (663,861)     & 0.0520 & 883451 & 8.8345 & Normal & -- & -- & -- \\
 & 1056 & -- & (861,1056)    & 0.0605 & 883922 & 8.8392 & Normal & -- & -- & -- \\
 & 1247 & -- & (1056,1247)   & 0.0415 & 883255 & 8.8326 & Normal & -- & -- & -- \\
 & 1442 & -- & (1247,1442)   & 0.0499 & 883198 & 8.8320 & Normal & -- & -- & -- \\
 & 1641 & -- & (1442,1641)   & 0.0535 & 883845 & 8.8384 & Normal & -- & -- & -- \\

\hline
\end{tabular}

\vspace{3pt}
{\footnotesize
\raggedright
1) Only complete ECG segments bounded by two detected R-peaks are evaluated. \\
2) Inference time was measured on a PSoC~6 MCU operating at 100~MHz. \\
3) ECG samples are derived from the MIT--BIH Arrhythmia Database and provided to the firmware as ADC-equivalent inputs. \\
4) All R-peak locations and segment boundaries are expressed in sample indices. \\
5) Annotation symbols follow the MIT--BIH convention: `N' denotes normal beats, and `/` denotes paced beats. \\
6) Live-captured ECG signals are acquired using the developed wearable patch and processed in real time on-device without external dataset input. \\
7) Baseline DNN autoencoder: normal if reconstruction error \(\leq 0.0816\)
\par}

\end{table*}

\paragraph{ECG Replay-Based Evaluation}

In this experiment, the ECG emulator-based evaluation framework discussed in Section~\ref{sec:evaluation_framework} is used to validate the firmware and TinyML inference on the wearable, using the hyperparameter-tuned DNN autoencoder. For the DNN autoencoder, five representative MIT--BIH records corresponding to normal and abnormal rhythms, namely Records~100, 106, 107, 118, and 207, were selected for analog replay using the ECG emulator. For each record, the first 3,600 samples of the ML-II channel, corresponding to 10~s of ECG data at a sampling rate of 360~samples/s, were replayed. The emulator output was connected directly to the ADC input of the PSoC~6, thereby bypassing the AD8232-based analog front-end. This configuration was used to avoid saturation or clipping of the emulator-generated ECG waveform, which has a relatively higher amplitude than the ECG signals acquired through the wearable electrodes. Table~\ref{tab:dnn_replay_results} presents the representative DNN autoencoder replay results.

\begin{table*}[!htbp]
\centering
\caption{ECG beat classification results and timing measurements from processing preloaded MIT--BIH ECG samples on the wearable.}
\label{tab:ecg_replay_validation}

\begin{tabular}{ccccccc}
\hline
\textbf{Record} &
\textbf{Representative} &
\textbf{MIT-BIH} &
\textbf{Autoencoder} &
\textbf{Multiclass} &
\textbf{Processing} &
\textbf{Average Inference} \\
&
\textbf{Segment} &
\textbf{Annotation} &
\textbf{Output} &
\textbf{Output} &
\textbf{Time (ms)} &
\textbf{Cycles (Time in ms)} \\
\hline

100 & (77,370)    & N & Normal   & N & 4.71 & 138064 (1.38) \\
107 & (268,565)   & / & Abnormal & / & 4.71 & 137976 (1.38) \\
214 & (347,661)   & L & Abnormal & L & 4.71 & 138182 (1.38) \\
232 & (1001,1660) & A & Abnormal & A & 4.71 & 138196 (1.38) \\
233 & (320,750)   & V & Abnormal & V & 4.71 & 138031 (1.38) \\

\hline
\end{tabular}
\end{table*}

\begin{table*}[!htbp]
\centering
\caption{ECG emulator-based evaluation of the autoencoder model on the wearable.}
\label{tab:dnn_replay_results}
\renewcommand{\arraystretch}{1.15}

\begin{tabular}{ccccccc}
\hline
\textbf{Record} &
\textbf{Segments} &
\textbf{Reconstruction-Error} &
\textbf{Mean Reconstruction} &
\textbf{Autoencoder} &
\textbf{Average Processing} &
\textbf{Average Inference} \\
&
\textbf{Processed} &
\textbf{Range} &
\textbf{Error} &
\textbf{Output} &
\textbf{Time (ms)} &
\textbf{Cycles / (ms)} \\
\hline

100 & 5 & 0.0352--0.0400 & 0.0381 & Normal   & 5.0575 & 394680 (3.95) \\
107 & 6 & 0.1247--0.2295 & 0.1813 & Abnormal & 5.0575 & 394173 (3.94) \\
207 & 4 & 0.1269--0.1754 & 0.1420 & Abnormal & 5.0575 & 394748 (3.95) \\
118 & 5 & 0.0821--0.0940 & 0.0884 & Abnormal & 5.0576 & 394694 (3.95) \\
106 & 4 & 0.0957--0.1443 & 0.1122 & Abnormal & 5.0577 & 395056 (3.95) \\

\hline
\end{tabular}
{\footnotesize 
\par\vspace{3pt}
* Hyperparameter-tuned DNN autoencoder: normal if reconstruction error \(\leq 0.0615\).\par}
\end{table*}

\section{Conclusion}
\label{sec_conclusion}

This work presented the development, optimization, and deployment of neural-network-based TinyML models for on-device ECG analysis on a custom-developed wearable.

The battery-operated wearable device, built around a PSoC~6 microcontroller, integrates ECG acquisition, signal conditioning, R--R interval-based segmentation, preprocessing, and neural-network inference for arrhythmia detection and multi-class ECG beat classification. By performing ECG analysis directly on the wearable device, the proposed approach minimizes the need to wirelessly transmit sensitive physiological data to external computing platforms for analysis. This helps protect patient data privacy while reducing the energy consumption associated with frequent wireless communication.

A systematic hyperparameter optimization approach was employed to identify compact autoencoder architectures suitable for embedded deployment. A scoring mechanism was developed to jointly account for detection performance, reconstruction-error thresholds, and model size during architecture selection. The results demonstrated that increasing neural-network complexity, and consequently model size, does not necessarily improve detection performance. Instead, careful selection of network architectures and hyperparameters can yield smaller models with lower inference latency, without compromising detection performance. This finding is particularly relevant to resource-constrained embedded platforms, where memory and computational resources are limited. In addition to arrhythmia detection, this work extended the analysis to multi-class ECG beat classification using a custom classifier derived from the hyperparameter-optimized autoencoder architecture.

The optimized models were quantized to INT8 and deployed on the PSoC~6 platform using TensorFlow Lite Micro. Following deployment, on-device inference was evaluated using both preloaded ECG segments and live ECG signals acquired from a subject. In addition, this work introduced an ECG emulator-based evaluation framework to enable extended testing of the wearable platform under controlled conditions. The emulator, developed using an ESP32-S3 microcontroller and a DAC-based waveform generation circuit, reproduces ECG waveforms from annotated MIT-BIH recordings and supplies them as analog signals to the input of the wearable. This setup emulates a subject by providing continuous ECG signals to the wearable device, enabling the evaluation of its hardware, firmware, and on-device inference pipeline over extended periods. 

Overall, this work extends our earlier ArrythML-I study~\cite{Nagarajan2026ArrythML} through systematic hyperparameter optimization of autoencoder architectures, six-class ECG beat classification, and the deployment and evaluation of the models on a custom-developed, battery-operated wearable device. Future work will focus on extending the evaluation to additional subjects and real-world wearable conditions, investigating dry-electrode-based wearable systems for long-term cardiac monitoring, and further optimizing power consumption to support longer-duration battery-operated operation.

\section*{Acknowledgment}

The authors acknowledge the M.Tech. students Chandra Shekhar Jha and Satya Prakash Chandra who provided engineering support for the development and integration of the wearable ECG hardware prototype~\cite{Nagarajan2026ECGSense}. The authors also acknowledge the support provided by MINRO, IIIT Bangalore, for carrying out this work.


The authors used AI-assisted tools (ChatGPT, Gemini) to improve the language, grammar, clarity, and readability of the manuscript.  All technical content, including the research methodology, experimental design, implementation, analysis, results, and conclusions, was developed and verified by the authors.

\sloppy
\bibliographystyle{IEEEtran}
\bibliography{myContents/references}

@inproceedings{vaidya2020gpio,
  title={GPIO PUF for IoT devices},
  author={Vaidya, Girish and Prabhakar, TV and Manjunath, Laveena},
  booktitle={GLOBECOM 2020-2020 IEEE Global Communications Conference},
  pages={1--6},
  year={2020},
  organization={IEEE}
}

@inproceedings{naz2024led,
  title={LED-PUF: Physical Unclonable LED Signatures For Unique Identification of IoT Nodes},
  author={Naz, Shab and Mulla, Ayesha Daniya and Sumayya, H and Sarang, Rifah Sameen and Polachan, Kurian},
  booktitle={2024 IEEE 22nd Mediterranean Electrotechnical Conference (MELECON)},
  pages={862--867},
  year={2024},
  organization={IEEE}
}

@inproceedings{chakrabarty2024iopuf,
  title={ioPUF: A PUF-based IoT Node Identification Utilizing Pull-Up/Down Resistors on IO Pins},
  author={Chakrabarty, Pralay and Ravi, Ananya Lakshmi and Polachan, Kurian},
  booktitle={2024 IEEE 22nd Mediterranean Electrotechnical Conference (MELECON)},
  pages={45--50},
  year={2024},
  organization={IEEE}
}

@inproceedings{agarwal2024vlc,
  title={VLC-LE: Energy-Efficient and Secure Visible Light Communication for the Internet of Bodies},
  author={Agarwal, Kavya and Lalwani, Ritesh and Abeena, U and Polachan, Kurian},
  booktitle={2024 International Conference on Smart Applications, Communications and Networking (SmartNets)},
  pages={1--6},
  year={2024},
  organization={IEEE}
}

@inproceedings{kambhampati2025murmur,
  title={Murmur: A Secure and Low-Energy Audio Communication for the Internet of Bodies},
  author={Kambhampati, Phani Pavan and Jagan, P. and Polachan, Kurian and Rao, Madhav},
  booktitle={2025 17th International Conference on COMmunication Systems and NETworks (COMSNETS)},
  pages={217--224},
  year={2025},
  organization={IEEE},
  doi={10.1109/COMSNETS63124.2025.10860573}
}

@inproceedings{majumdar2025hbclock,
  title={hbcLock: Encrypted RF Communication Utilizing Body-Coupled Keys for the Internet of Bodies},
  author={Majumdar, Soumick and Madurwar, Anshul and Shetty, Anmol and Polachan, Kurian},
  booktitle={2025 38th International Conference on VLSI Design and 2024 23rd International Conference on Embedded Systems (VLSID)},
  pages={55--60},
  year={2025},
  organization={IEEE}
}

@article{Moody2001,
  author    = {G. B. Moody and R. G. Mark},
  title     = {The Impact of the MIT-BIH Arrhythmia Database},
  journal   = {IEEE Engineering in Medicine and Biology Magazine},
  volume    = {20},
  number    = {3},
  pages     = {45--50},
  year      = {2001},
  month     = {May-Jun},
  doi       = {10.1109/51.932724}
}

@book{Rangayyan2002,
  author    = {Rangaraj M. Rangayyan},
  title     = {Biomedical Signal Analysis: A Case-Study Approach},
  publisher = {IEEE Press and John Wiley \& Sons},
  year      = {2002},
  address   = {New York, NY, USA}
}

@misc{MITBIHDatabase,
  author       = {Moody, George B. and Mark, Roger G.},
  title        = {The MIT-BIH Arrhythmia Database},
  year         = {1992},
  url          = {https://physionet.org/content/mitdb/1.0.0/},
  note         = {Available at PhysioNet},
  howpublished = {PhysioNet},
  doi          = {10.13026/C2F305},
}

@article{kakria2015realtime,
  author    = {Kakria, P. and Tripathi, N. K. and Kitipawang, P.},
  title     = {A Real-Time Health Monitoring System for Remote Cardiac Patients Using Smartphone and Wearable Sensors},
  journal   = {International Journal of Telemedicine and Applications},
  volume    = {2015},
  pages     = {373474},
  year      = {2015},
  doi       = {10.1155/2015/373474},
  note      = {Epub 2015 Dec 15. PMID: 26788055; PMCID: PMC4692989}
}

@article{Heydari2025TinyMLSurvey,
  author  = {Soroush Heydari and Qusay H. Mahmoud},
  title   = {Tiny Machine Learning and On-Device Inference: A Survey of Applications, Challenges, and Future Directions},
  journal = {Sensors},
  volume  = {25},
  pages   = {3191},
  year    = {2025},
  doi     = {10.3390/s25103191}
}

@article{elhanashi2024advancements,
  AUTHOR = {Elhanashi, Abdussalam and Dini, Pierpaolo and Saponara, Sergio and Zheng, Qinghe},
  TITLE = {Advancements in TinyML: Applications, Limitations, and Impact on IoT Devices},
  JOURNAL = {Electronics},
  VOLUME = {13},
  YEAR = {2024},
  NUMBER = {17},
  ARTICLE-NUMBER = {3562},
  URL = {https://www.mdpi.com/2079-9292/13/17/3562},
  ISSN = {2079-9292},
  DOI = {10.3390/electronics13173562}
}

@article{Kim2023TinyMLECG,
  author  = {Eunchan Kim and Jaehyuk Kim and Juyoung Park and Haneul Ko and Yeunwoong Kyung},
  title   = {TinyML-Based Classification in an ECG Monitoring Embedded System},
  journal = {Computers, Materials \& Continua},
  volume  = {75},
  number  = {1},
  pages   = {1751--1764},
  year    = {2023},
  doi     = {10.32604/cmc.2023.031663}
}

@article{Hizem2025ECGEdge,
  author  = {Moez Hizem and Leila Bousbia and Yassmine Ben Dhiab and Mohamed Ould-Elhassen Aoueileyine and Ridha Bouallegue},
  title   = {Reliable ECG Anomaly Detection on Edge Devices for Internet of Medical Things Applications},
  journal = {Sensors},
  volume  = {25},
  number  = {8},
  pages   = {2496},
  year    = {2025},
  doi     = {10.3390/s25082496}
}

@article{Akbari2025AFEdge,
  author  = {Yash Akbari and Ningrong Lei and Nilesh Patel and Yonghong Peng and Oliver Faust},
  title   = {Atrial Fibrillation Detection on the Embedded Edge: Energy-Efficient Inference on a Low-Power Microcontroller},
  journal = {Sensors},
  volume  = {25},
  number  = {21},
  pages   = {6601},
  year    = {2025},
  doi     = {10.3390/s25216601}
}

@inproceedings{Nagarajan2025Autoencoder,
  author    = {Nagarajan, S. and Polachan, Kurian},
  title     = {An Autoencoder-Based TinyML Model for On-Device Arrhythmia Detection},
  booktitle = {2025 International Conference on Smart Applications, Communications and Networking (SmartNets)},
  pages     = {1--6},
  year      = {2025},
  publisher = {IEEE},
  doi       = {10.1109/SmartNets65254.2025.11106907}
}

@misc{psoc6_proto_ble,
  author       = {{Infineon Technologies}},
  title        = {{CY8CPROTO-063-BLE PSoC™ 6 BLE Prototyping Kit}},
  howpublished = {\url{https://www.infineon.com/evaluation-board/CY8CPROTO-063-BLE}},
  year         = {2023},
  note         = {Accessed: Jan. 2026}
}

@misc{ad8232_datasheet,
  author       = {{Analog Devices, Inc.}},
  title        = {{AD8232: Single-Lead, Heart Rate Monitor Analog Front End}},
  howpublished = {\url{https://www.analog.com/en/products/ad8232.html}},
  year         = {2021},
  note         = {Accessed: Jan. 2026}
}

@book{warden2019tinyml,
  author    = {Warden, Pete and Situnayake, Daniel},
  title     = {TinyML: Machine Learning with TensorFlow Lite on Arduino and Ultra-Low-Power Microcontrollers},
  publisher = {O'Reilly Media},
  year      = {2019}
}

@misc{InfineonModusToolboxSoftware,
  title        = {ModusToolbox\texttrademark{} Software},
  author       = {{Infineon Technologies AG}},
  year         = {2025},
  note         = {Available: https://www.infineon.com/design-resources/development-tools/sdk/modustoolbox-software}
}

@inproceedings{clark2018wearable,
  author    = {Clark, Nicholar and Sandor, Edward and Walden, Calvin and Ahn, In Soo and Lu, Yufeng},
  title     = {A Wearable ECG Monitoring System for Real-Time Arrhythmia Detection},
  booktitle = {2018 IEEE 61st International Midwest Symposium on Circuits and Systems (MWSCAS)},
  year      = {2018},
  pages     = {787--790},
  address   = {Windsor, ON, Canada},
  publisher = {IEEE},
  doi       = {10.1109/MWSCAS.2018.8624097}
}

@article{alimbayeva2024wearable,
  title={Wearable {ECG} Device and Machine Learning for Heart Monitoring},
  author={Alimbayeva, Zhadyra and Alimbayev, Chingiz and Ozhikenov, Kassymbek and Bayanbay, Nurlan and Ozhikenova, Aiman},
  journal={Sensors},
  volume={24},
  number={13},
  pages={4201},
  year={2024},
  publisher={MDPI},
  doi={10.3390/s24134201}
}

@misc{das2019heartbeatclassificationwearablesusing,
title={Heartbeat Classification in Wearables Using Multi-layer Perceptron and Time-Frequency Joint Distribution of ECG},
author={Das, Anup and Catthoor, Francky and Schaafsma, Siebren},
year={2019},
eprint={1908.06865},
archivePrefix={arXiv},
primaryClass={eess.SP},
url={https://arxiv.org/abs/1908.06865}
}

@inproceedings{ingolfsson2021ecgtcn,
  title={{ECG-TCN}: Wearable Cardiac Arrhythmia Detection with a Temporal Convolutional Network},
  author={Ingolfsson, Thorir Mar and Wang, Xiaying and Hersche, Michael and Burrello, Alessio and Cavigelli, Lukas and Benini, Luca},
  booktitle={2021 IEEE 3rd International Conference on Artificial Intelligence Circuits and Systems (AICAS)},
  pages={1--4},
  year={2021},
  organization={IEEE},
  doi={10.1109/AICAS51828.2021.9458498}
}

@article{busia2024tinytransformer,
  author={Busia, Paola and Scrugli, Matteo Antonio and Jung, Victor Jean-Baptiste and Benini, Luca and Meloni, Paolo},
  journal={IEEE Transactions on Biomedical Circuits and Systems}, 
  title={A Tiny Transformer for Low-Power Arrhythmia classification on Microcontrollers}, 
  year={2025},
  volume={19},
  number={1},
  pages={142-152}, 
  doi={10.1109/TBCAS.2024.3401858}
}

@article{ecgcnnclassification,
  author    = {Kiranyaz, Serkan and Ince, Turker and Gabbouj, Moncef},
  title     = {Real-Time Patient-Specific ECG Classification by 1-D Convolutional Neural Networks},
  journal   = {IEEE Transactions on Biomedical Engineering},
  volume    = {63},
  number    = {3},
  pages     = {664--675},
  year      = {2016},
  publisher = {IEEE},
  doi       = {10.1109/TBME.2015.2468589}
}

@inproceedings{Nagarajan2026ECGSense,
  author    = {Nagarajan, S. and Jha, Chandra Shekhar and Chandra, Satya Prakash and Polachan, Kurian},
  title     = {{ECG-Sense}: Demonstration of a Wearable {ECG} Patch with an On-Device {TinyML} Engine for Real-Time Arrhythmia Detection},
  booktitle = {2026 International Conference on Smart Applications, Communications and Networking (SmartNets)},
  year      = {2026},
  pages     = {1--6},
  doi       = {10.1109/SmartNets69662.2026.11604743}
}

@Article{s25092848,
AUTHOR = {Abdelrazik, Ahmed and Eldesouky, Mahmoud and Antoun, Ibrahim and Lau, Edward Y. M. and Koya, Abdulmalik and Vali, Zakariyya and Suleman, Safiyyah A. and Donaldson, James and Ng, G. André},
TITLE = {Wearable Devices for Arrhythmia Detection: Advancements and Clinical Implications},
JOURNAL = {Sensors},
VOLUME = {25},
YEAR = {2025},
NUMBER = {9},
ARTICLE-NUMBER = {2848},
URL = {https://www.mdpi.com/1424-8220/25/9/2848},
PubMedID = {40363284},
ISSN = {1424-8220},
DOI = {10.3390/s25092848}
}

@misc{vitalconnect2026vitalpatch,
  author       = {{VitalConnect}},
  title        = {{VitalPatch}: Frequently Asked Questions},
  howpublished = {[Online]},
  url          = {https://vitalconnect.com/patients/faqs/},
  note         = {Accessed: Nov. 13, 2026}
}

@manual{irhythm2025zioxt,
  author       = {{iRhythm Technologies, Inc.}},
  title        = {{Zio XT Clinical Reference Manual}},
  organization = {iRhythm Technologies, Inc.},
  year         = {2025},
  url          = {https://www.irhythmtech.com/},
  note         = {Accessed: Nov. 13, 2026}
}

@article{rajbhandary2022ecg,
  author  = {Rajbhandary, Paurakh Lal and Nallathambi, Gabriel
             and Selvaraj, Nandakumar and Tran, Thang
             and Colliou, Olivier},
  title   = {{ECG} Signal Quality Assessments of a Small
             Bipolar Single-Lead Wearable Patch Sensor},
  journal = {Cardiovascular Engineering and Technology},
  volume  = {13},
  number  = {5},
  pages   = {783--796},
  year    = {2022},
  doi     = {10.1007/s13239-022-00617-3}
}

@article{niu2023diagnostic,
  author  = {Niu, Yonghong and Wang, Hao and Wang, Hong
             and Zhang, Hui and Jin, Zhigeng and Guo, Yutao},
  title   = {Diagnostic Validation of Smart Wearable Device
             Embedded with Single-Lead Electrocardiogram
             for Arrhythmia Detection},
  journal = {Digital Health},
  volume  = {9},
  pages   = {20552076231198682},
  year    = {2023},
  doi     = {10.1177/20552076231198682}
}

@article{kwon2024telemonitoring,
  author  = {Kwon, Soonil and Choi, Eue-Keun
             and Lee, So-Ryoung and Oh, Seil
             and Song, Hee-Seok and Lee, Young-Shin
             and Han, Sang-Jin and Lim, Hong Euy},
  title   = {Comparison of Novel Telemonitoring System
             Using the Single-lead Electrocardiogram
             Patch With Conventional Telemetry System},
  journal = {Korean Circulation Journal},
  volume  = {54},
  number  = {3},
  pages   = {140--153},
  year    = {2024},
  doi     = {10.4070/kcj.2023.0252}
}

@inproceedings{abdou2024accelerator,
  author    = {Abdou, Abdelrahman and Krishnan, Sridhar},
  title     = {Hardware Accelerator for a Power Efficient
               Single-lead Dry-electrode {ECG} Wearable Design},
  booktitle = {2024 46th Annual International Conference
               of the IEEE Engineering in Medicine
               and Biology Society (EMBC)},
  pages     = {1--4},
  year      = {2024},
  doi       = {10.1109/EMBC53108.2024.10782919}
}

@article{choi2024exercise,
  author  = {Choi, Hyo-In and Lee, Seung Jae
             and Choi, Jong Doo and Kim, GyungChul
             and Lee, Young-Shin and Lee, Jong-Young},
  title   = {Efficacy of Wearable Single-Lead {ECG}
             Monitoring during Exercise Stress Testing:
             A Comparative Study},
  journal = {Sensors},
  volume  = {24},
  number  = {19},
  pages   = {6394},
  year    = {2024},
  doi     = {10.3390/s24196394}
}

@article{shin2026comparative,
  author  = {Shin, Dong Geum and Kim, Bokyoung
             and Ahn, Jihyun and Park, Hyung Wook
             and Yoon, Namsik and Lee, Kihong
             and Kim, Yoo Ri},
  title   = {Comparative Performance of Patch-Type
             and Lead-Type Wearable Electrocardiogram
             Devices for Arrhythmia Detection
             in Routine Clinical Practice},
  journal = {Journal of Clinical Medicine},
  volume  = {15},
  number  = {2},
  pages   = {526},
  year    = {2026},
  doi     = {10.3390/jcm15020526}
}

@article{Nagarajan2026ArrythML,
  author  = {Nagarajan, S. and Polachan, Kurian},
  title   = {{ArrythML}: An Autoencoder-Based {TinyML} Approach for
             On-Device Arrhythmia Detection on Resource-Constrained
             Embedded Systems},
  journal = {IEEE Access},
  year    = {2026},
  doi     = {10.1109/ACCESS.2026.3751785}
}

@inproceedings{porlapothula2025cappuf,
  title={{CapPUF}: A Tamper-Resistant {PUF} Utilizing {I/O} Capacitances for {IoT} Nodes},
  author={Porlapothula, Dilli Babu and Polachan, Kurian},
  booktitle={2025 IEEE International Symposium on Circuits and Systems (ISCAS)},
  pages={1--5},
  year={2025},
  organization={IEEE}
}

@article{baig2017systematic,
  author  = {Baig, Mirza Mansoor and GholamHosseini, Hamid and
             Moqeem, Aasia A. and Mirza, Farhaan and
             Lind{\'e}n, Maria},
  title   = {A Systematic Review of Wearable Patient Monitoring Systems---Current Challenges and Opportunities for Clinical Adoption},
  journal = {Journal of Medical Systems},
  year    = {2017},
  volume  = {41},
  number  = {7},
  pages   = {115},
  doi     = {10.1007/s10916-017-0760-1}
}

\begin{IEEEbiography}
[{\includegraphics[width=1in,height=1.25in,clip,keepaspectratio]{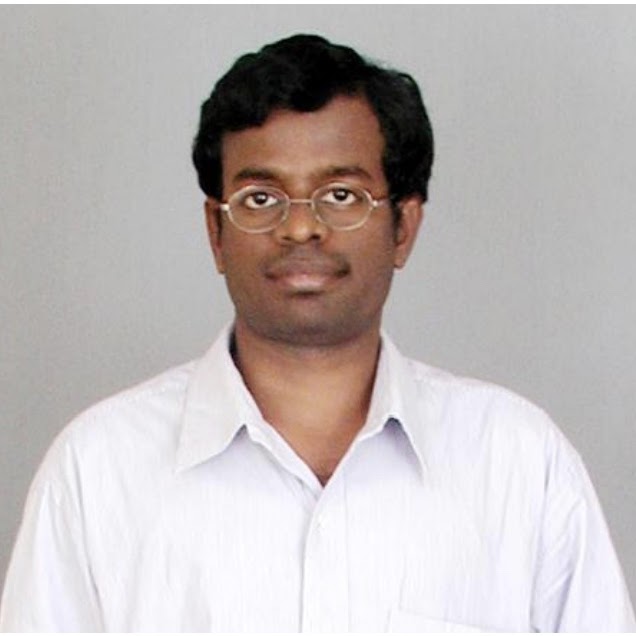}}]{Nagarajan S}
received the M.Tech. degree in Embedded Systems and has over 20 years of industry experience in embedded systems and firmware development. He is currently a doctoral researcher at the International Institute of Information Technology Bangalore (IIIT-B), India. His research interests include TinyML, embedded machine learning, real-time embedded systems, real-time signal processing, and edge AI for wearable healthcare systems.
\end{IEEEbiography}


\begin{IEEEbiography}
[{\includegraphics[width=1in,height=1.25in,clip,keepaspectratio]{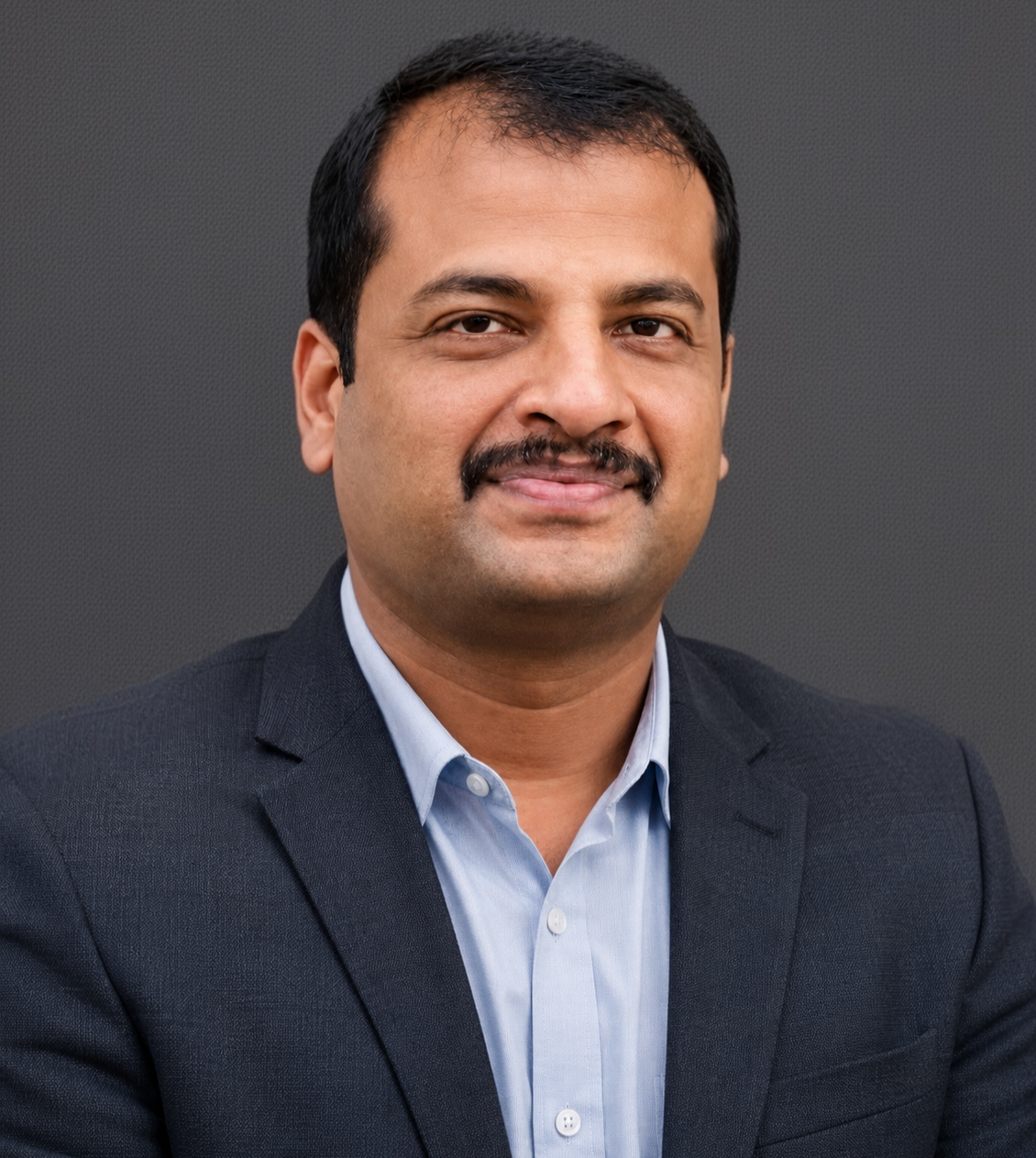}}]{Kurian Polachan}
(Senior Member, IEEE) is an Assistant Professor with the International Institute of Information Technology Bangalore (IIIT-B), India, where he leads the Connected Devices and Wearables Lab (CDWL). He received the M.Tech. and Ph.D. degrees from the Indian Institute of Science (IISc), Bangalore, India. From 2021 to 2022, he was a Postdoctoral Researcher with Purdue University, West Lafayette, IN, USA. Before joining academia, he spent over five years with Cypress Semiconductor, focusing on embedded systems design. His research interests include secure and intelligent wearable systems for next-generation healthcare and IoT applications, with a specific focus on ultra-low-power hardware, embedded intelligence, and hardware-rooted security.
\end{IEEEbiography}

\end{document}